\documentclass[final,1p,times]{elsarticle}
\usepackage{amsmath,amssymb,amsfonts}

\usepackage{algorithmic}
\usepackage{graphicx}
\usepackage{textcomp}
\usepackage{xcolor}
\usepackage{pifont}
\usepackage{tikz}
\usetikzlibrary{positioning}
\usetikzlibrary{shapes.geometric}
\usepackage{multirow}
\usetikzlibrary{calc}
\begin{document}
\begin{frontmatter}
\title{Increasing Depth of Neural Networks for Life-long Learning}

% \author{\IEEEauthorblockN{1\textsuperscript{st} Jedrzej Kozal}
% \IEEEauthorblockA{\textit{Department of Systems and Computer Networks} \\
% \textit{Wrocław University of Science and Technology}\\
% Wrocław, Poland \\
% jedrzej.kozal@pwr.edu.pl}
% \and
% \IEEEauthorblockN{2\textsuperscript{nd} Michał Wozniak}
% \IEEEauthorblockA{\textit{Department of Systems and Computer Networks} \\
% \textit{Wrocław University of Science and Technology}\\
% Wrocław, Poland \\
% michal.wozniak@pwr.edu.pl}
% }

% \tnotetext[t1]{This document is the results of the research project funded by the National Science Foundation.}

\author{Jedrzej Kozal\corref{cor1}$^{[0000-0001-7336-2561]}$ and Michal Wozniak$^{[0000-0003-0146-4205]}$}

\ead{jedrzej.kozal@pwr.edu.pl,michal.wozniak@pwr.edu.pl}

\address{Department of Systems and Computer Networks,\\ 
    Faculty of Information and Communication Technology,\\ 
    Wroclaw University of Science and Technology,\\
    Wybrzeze Wyspianskiego 27, 50-370 Wroclaw, Poland\\}
    
\cortext[cor1]{Corresponding author}

\begin{abstract}
%Increasing neural network depth is a well-known method for improving neural network performance. This work explores whether extending neural network depth may be beneficial in a life-long learning setting. We propose a novel method based on adding new layers on top of existing ones to enable the forward transfer of knowledge and adapting previously learned representations. We utilize a method of determining the most similar tasks for selecting the best location in our network to add new nodes with trainable parameters. This approach allows for creating a tree-like model, where each node is a set of neural network parameters dedicated to a specific task. 
%The proposed method is inspired by Progressive Neural Network concept. Therefore it benefits from dynamic changes in network structure. It is compatible with commonly used computer vision architectures and does not require a custom network structure. As an adaptation to changing data distribution is made by expanding the architecture, there is no need to utilize a rehearsal buffer. For this reason, our method can be used for sensitive applications where data privacy must be considered.
%Experiments on Split CIFAR and Split Tiny ImageNet show that the proposed algorithm is on par with other continual learning methods.
\noindent\textbf{Purpose:} We propose a novel method for continual learning based on the increasing depth of neural networks. This work explores whether extending neural network depth may be beneficial in a life-long learning setting. 

\noindent \textbf{Methods:} We propose a novel approach based on adding new layers on top of existing ones to enable the forward transfer of knowledge and adapting previously learned representations. We employ a method of determining the most similar tasks for selecting the best location in our network to add new nodes with trainable parameters. This approach allows for creating a tree-like model, where each node is a set of neural network parameters dedicated to a specific task. The Progressive Neural Network concept inspires the proposed method. Therefore, it benefits from dynamic changes in network structure. 
However, Progressive Neural Network allocates a lot of memory for the whole network structure during the learning process. The proposed method alleviates this by adding only part of a network for a new task and utilizing a subset of previously trained weights. At the same time, we may retain the benefit of PNN, such as no forgetting guaranteed by design, without needing a memory buffer.

\noindent \textbf{Results:} Experiments on Split CIFAR and Split Tiny ImageNet show that the proposed algorithm is on par with other continual learning methods. In a more challenging setup with a single computer vision dataset as a separate task, our method outperforms Experience Replay.

\noindent \textbf{Conclusion:} It is compatible with commonly used computer vision architectures and does not require a custom network structure. As an adaptation to changing data distribution is made by expanding the architecture, there is no need to utilize a rehearsal buffer. For this reason, our method could be used for sensitive applications where data privacy must be considered.

\end{abstract}


\end{frontmatter}

%\linenumber

%\maketitle

\section{Introduction}

Neural networks may benefit greatly from increasing depth \cite{DBLP:journals/corr/HeZRS15,DBLP:journals/corr/HuangSLSW16,DBLP:journals/corr/abs-1905-11946}. Especially in computer vision introduction of residual connection \cite{DBLP:journals/corr/HeZRS15} allowed for training much deeper networks, which further boosted the results obtained on ILSVRC ImageNet challenge \cite{DBLP:journals/corr/RussakovskyDSKSMHKKBBF14}, stimulating further research based on this architecture \cite{DBLP:journals/corr/XieGDTH16,DBLP:journals/corr/SzegedyIV16,DBLP:journals/corr/abs-1905-11946}. %Greater depth increases the model's capacity and allows for learning of more complicated functions. 
Greater depth increases models' capacity and allows for learning more complicated concepts. 
Features learned by convolutional network are hierarchical \cite{DBLP:journals/corr/ZeilerF13}. This could be shown with techniques that allow for the visualization of filters learned by the network. Filters at the network's bottom contain generic features responsible for detecting edges or color conjunctions. With increasing depth, filters in convolutional layers are responsible for recognizing more abstract concepts. Layers at the top of a network are specific to the task solved at hand. This property is used in the transfer learning technique \cite{Bozinovski2020ReminderOT,DBLP:journals/corr/abs-1912-11370}, where the model pre-trained on a large dataset is later fine-tuned on a typically smaller dataset. It helps avoid collecting a large dataset, which could be prohibitive for some applications. 
Fine-tuning of already pretrained backbones was also utilized in life-long learning \cite{DBLP:journals/corr/LiH16e}, as it allows transmitting learned knowledge among different tasks. 

Canonical machine learning methods assume i.d.d. data, while life-long learning or continual learning should consider that this assumption is not met. Especially in life-long learning, we assume that the new data arrive sequentially and new incoming data could be sampled from a different distribution than the previous ones.  
The arrival of new concepts to generalize leads to forgetting already learned concepts. This phenomenon is known as \emph{catastrophic forgetting} \cite{catastrophic_forgetting}.
For this reason, continual learning algorithms should be able to adapt to changes in data distribution and simultaneously contain mechanisms that prevent the forgetting of already acquired knowledge \cite{Chen:2018}. In literature this notion is known as \emph{stability-plasticity dilemma} \cite{7296710}. A similar concept was captured as a generalization-forgetting trade-off in \cite{DBLP:journals/corr/abs-2109-14035}. Both humans and animals can learn new tasks without degrading already acquired knowledge.  
Our brains handle continual learning well and can refine knowledge based on new experiences to make better decisions \cite{human_learning}. For this reason, life-long learning is closer to human learning than standard machine learning. Also, continual learning development is essential for building autonomous learning agents and robots capable of operating in the open world. If no closed set of knowledge could be defined, then the agent must learn constantly based on experiences. 
Perpetual learning from scratch is largely ineffective in this setup as a model would have been trained too often. This has sparked interest in recent years in continual learning \cite{DBLP:journals/corr/abs-1802-07569}.

This work combines observations made by the deep learning community over the years in neural network architecture development and life-long learning. We evaluate the ability to increase network depth for life-long learning, stacking new layers on top of existing ones. 
Using an approach based on increasing depth, we hope to benefit from increasing the depth and training of new parameters by utilizing previously learned representations. This allows for direct forward knowledge transfer that emerges as a natural consequence of the proposed method.

%In summary our contributions are following:
In summary, the contributions of our work are as follows:
\begin{enumerate}
    \item Proposing a new rehearsal-free method based on the increasing depth of neural network architecture. % According to our best knowledge, this is the first method that directly utilizes the benefits of increasing depth for continual learning.
    \item Conducting a thorough experimental evaluation with benchmarks commonly used in continual learning and a more demanding multi-dataset environment. 
    \item Performing in-depth ablation studies that evaluate the impact of each part of the proposed method.
%    \item Discussion of obtained results was carried out with references and comparison to existing works in literature.
\end{enumerate}

\section{Related Works}

Increasing the depth of neural network architectures is an inherent part of deep learning progress. Most breakthroughs in computer visions may be attributed to increasing the size of network \cite{vgg,DBLP:journals/corr/SzegedyVISW15,DBLP:journals/corr/HeZRS15,DBLP:journals/corr/abs-1905-11946,DBLP:journals/corr/abs-2104-00298}. With this progress, other areas of computer vision started to benefit from the utilization of large pretrained backbones \cite{DBLP:journals/corr/GirshickDDM13,DBLP:journals/corr/HeGDG17}. In \cite{DBLP:journals/corr/HuangSLSW16} %One interesting proposal was 
Stochastic Depth %, which 
was introduced to train very deep networks, allowing ResNets with depths exceeding 1000 layers to be trained. It was later incorporated into newer architectures \cite{DBLP:journals/corr/abs-1905-11946,DBLP:journals/corr/abs-2104-00298}. 
Over time Neural Architecture Search (NAS) algorithms have been used to find new neural network architectures \cite{DBLP:journals/corr/ZophL16, DBLP:journals/corr/ZophVSL17}. Due to this development, some researchers focused on determining the proper search space for NAS algorithms. In \cite{DBLP:journals/corr/abs-1905-11946} principles of expanding neural networks are studied in depth. The authors considered three dimensions for increasing network size: depth (number of layers), width (number of filters), and input resolution. This allowed for constructing a network with a smaller number of parameters and better performance.

Many algorithms have been developed so far for life-long learning. In \cite{DBLP:journals/corr/abs-1802-07569}, existing approaches are divided into three groups, namely regularization approaches,  dynamic architectures, and memory replay methods. In the following paragraphs, we will describe each of these groups.

\paragraph{Regularization methods}
The first group consists of methods that utilize the same parameters for training while employing some regularization factor in the learning process to alleviate forgetting. Elastic Weight Consolidation (EWC) \cite{DBLP:journals/corr/KirkpatrickPRVD16} is a method where a new regularization term was introduced to slow down learning for parameters that were important in previous tasks. 
Li and Hoiem \cite{DBLP:journals/corr/LiH16e} proposed Learning without Forgetting (LwF) - a method where a new classification head is added for each task, and a backbone is common for all tasks. Firstly, incoming data for the new task is labeled according to the outputs of classification heads for previous tasks. Then, these soft labels are used to prevent forgetting of earlier tasks. 

\paragraph{Dynamic architectures}
Expanding network capacity to handle changing data distribution is well-known in the literature. For example, progressive Neural Networks (PNN) \cite{DBLP:journals/corr/RusuRDSKKPH16} is a life-long learning method that introduces growing neural networks in terms of a number of parameters with each new task. When data for a new task arrives, a new column with randomly initialized weights is added to an existing structure. Each newly added layer takes, as an input, features generated by layers from current and previous tasks. For this reason, features learned in each layer are a function of data from the current task and knowledge learned from earlier tasks.
In \cite{DBLP:journals/corr/abs-1708-01547} extension of PNN was proposed, where expanding neural network is combined with selective retraining. Firstly network fine-tunes selected parameters that have already been learned. New layers are added during training if the resulting loss is greater than the defined threshold. L1 normalization is used to introduce sparsity into the network.
In \cite{DBLP:journals/corr/abs-1904-00310} NAS framework is utilized to modify network structure for life-long learning. Each layer is reused for every new task with no changes, fine-tuned, or expanded with new parameters.

\paragraph{Rehearsal-based methods}

Rehearsal methods \cite{DBLP:journals/corr/abs-1902-10486} create a memory buffer for storing and utilizing samples from previous tasks during learning new tasks.
Gradient Episodic Memory (GEM) \cite{DBLP:journals/corr/Lopez-PazR17} utilizes samples stored in memory to project gradients computed for the current task to ensure that loss over all previous tasks is not increasing. A-GEM \cite{DBLP:journals/corr/abs-1812-00420} is a modification of GEM that relaxes the condition of decreasing loss for previous tasks only to decrease the average loss of earlier tasks. This modification speeds up training with an approximate ability to solve the tasks.

\begin{figure*}[]
\begin{center}
    % \resizebox{330pt}{230pt}{
    %     \input{images/structure_schema}
    % }
    \includegraphics[width=13cm]{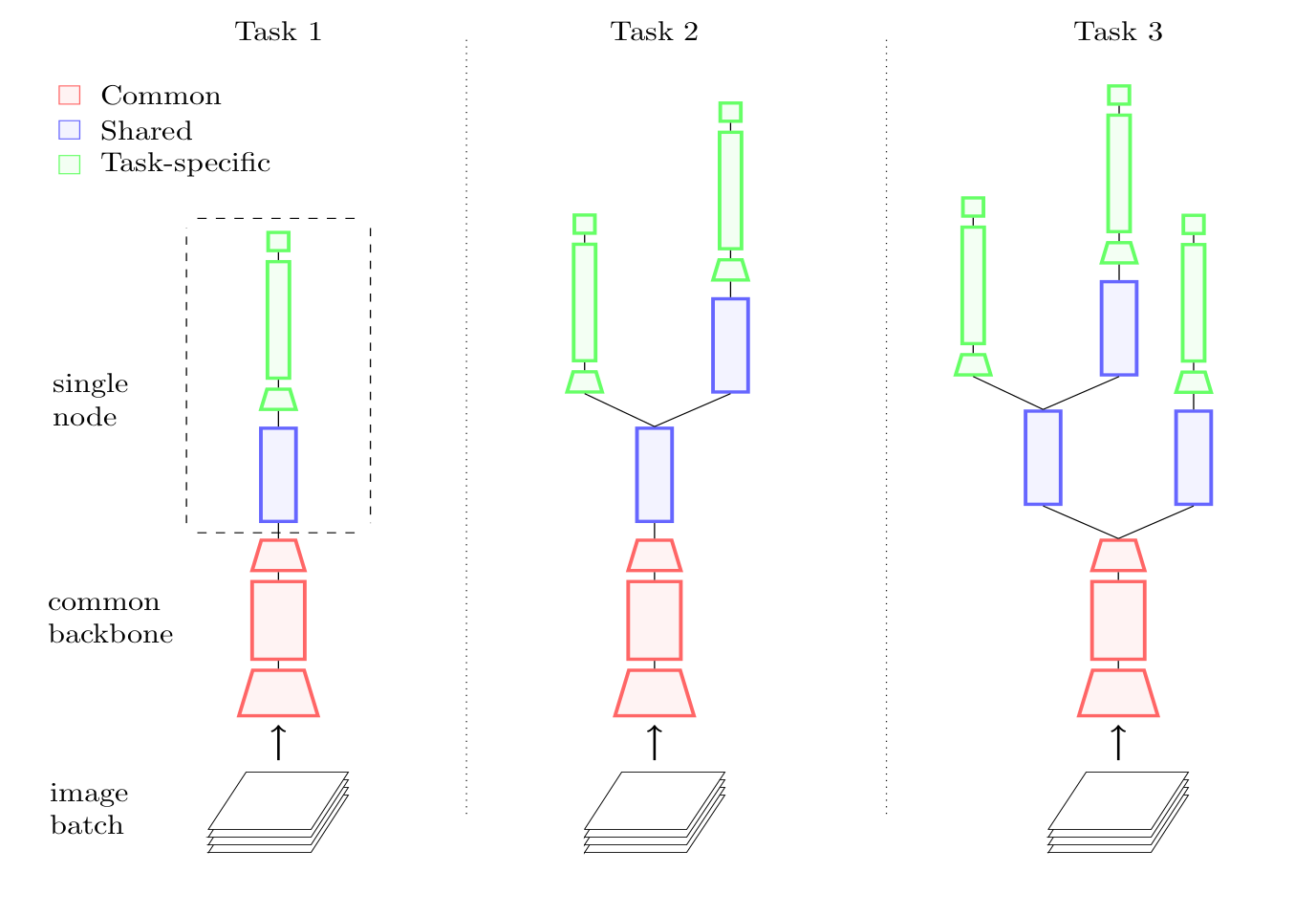}
\end{center}
\caption{Stacking layers for new tasks. Evolution of network structure corresponds to operations in Eq.~\ref{eqn:task_1_structure}, \ref{eqn:task_2_structure}, \ref{eqn:task_3_structure}. Rectangles denote convolutional layers, trapeze convolutions with stride 2 or pooling operations, and wide rectangles at the end of the network denote classifiers. Layers are divided into 3 groups: a common backbone for all tasks, common layers to which future tasks could be attached, and task-specific layers with separate classifiers. (Left) Standard neural network structure. The node that expands the network structure consists of the last two network blocks. (Middle) We add new layers by attaching them to previously shared weights. (Right) If the task similarity between the current and previously learned tasks is too small, we may attach a new node directly to the backbone.}
\label{schema}
\end{figure*}

\section{Material and methods}

This section provides a detailed description of the proposed algorithm along with the experiment setup. We first introduce problem setup for task-incremental learning in \ref{problem_setup} and proposed method in \ref{increasing_depth}. We describe the method for determining task similarity in \ref{task_similarity} and discuss the time complexity of the proposed method in \ref{time_complexity}. Known challenges related to training very deep networks are described in \ref{very_deep_neural_networks}. We discuss using different normalization methods in continual learning in \ref{batch_normalization} and provide a formula for model inference at test time in \ref{model_inference}.

\subsection{Problem Setup} \label{problem_setup}

Firstly, let us introduce a task incremental learning problem. In contrast to standard training, the stationary dataset model is trained with a sequence of $T$ tasks. Each task at step $t$ corresponds to the single dataset $\mathcal{D}_t = \{ ( x_{i}^t, y_{i}^{t}, t_i ) \}_{i=1}^{N_t}$ with model input $x_{i}^t$, label $y_{i}^{t} \in \mathcal{Y}_t$, where $\mathcal{Y}_t$ is the set of classes for the current task, and the task index $t_i$. We assume that the model cannot see the same classes twice, i.e., 
% $\forall_{t<T} \mathcal{Y}_t \cap \mathcal{Y}_{t+1} = \varnothing$ 
$\forall_{s,t\in\{1,...,T\}} Y_t \cap Y_s=\varnothing$ for $s\neq t$. 
After training with $\mathcal{Y}_t$ model cannot reaccess the data. The goal is to train the predictive model $p(y|x_i) = \mathcal{F}^{t}(x_i)$ for each task $t$ and learn all $T$ tasks sequentially without forgetting the knowledge essential for solving previous tasks.

\subsection{Increasing Depth} \label{increasing_depth}

We propose gradually increasing neural network depth to benefit from refined representations learned for previous tasks. 
We divide layers into three groups. First layers with common weights for all tasks (common backbone) denoted by $\mathcal{F}_B$. They are responsible for extracting basic general-purpose features utilized by all tasks. A backbone may contain pre-trained weights from ImageNet. Next, intermediate layers are shared by current and future tasks, denoted by $\mathcal{F}_S$. Activations produced by these layers are more task-specific and could be shared by later tasks if a future task is sufficiently similar to the current one. Lastly, layers specific to the current task are denoted by $\mathcal{F}_C$. These layers produce the most task-specific activations used only to classify the current task. The entire network structure with all three categories is shown in Fig.~\ref{schema}. Activations produced by each of these groups are input to the next group. With this notation standard network structure used for the first task may be written as:

\begin{equation}
    \label{eqn:task_1_structure}
    \mathcal{F}^1(.) = \mathcal{F}^1_C \circ \mathcal{F}^1_S \circ \mathcal{F}_B(.)
\end{equation}

\noindent where $\circ$ is the function composition operator.
When a new task $t$ is presented, we add new layers on top of already trained ones. This could be accomplished only if the tensor size of input and output activations of $\mathcal{F}_S$ is the same, so choosing the proper scope of shared layers is crucial for this method. If this assumption is met, we may create new common and task-specific layers and add them on top of already trained layers:

\begin{equation}
    \label{eqn:task_2_structure}
    \mathcal{F}^2(.) = \mathcal{F}^2_C\circ \mathcal{F}^2_S \circ \mathcal{F}^1_S \circ \mathcal{F}_B(.)
\end{equation}

\noindent Weights from previous tasks (including previously added common layers) and backbone are frozen during training. Only new layers added for the current task are trained. Due to this design, we may obtain simultaneous full knowledge retention and forward knowledge transfer. We will refer to shared and task-specific layers for a single task as a node. When a new task is dissimilar to all previous tasks, then existing representations should not be used for training. In this case, new layers could be attached to the backbone directly:

\begin{equation}
    \label{eqn:task_3_structure}
    \mathcal{F}^3(.) = \mathcal{F}^3_C \circ \mathcal{F}^3_S \circ \mathcal{F}_B(.)
\end{equation}

\begin{figure}[t]
\begin{center}
% \centerline{\input{images/example_structure}}
\includegraphics[width=5cm]{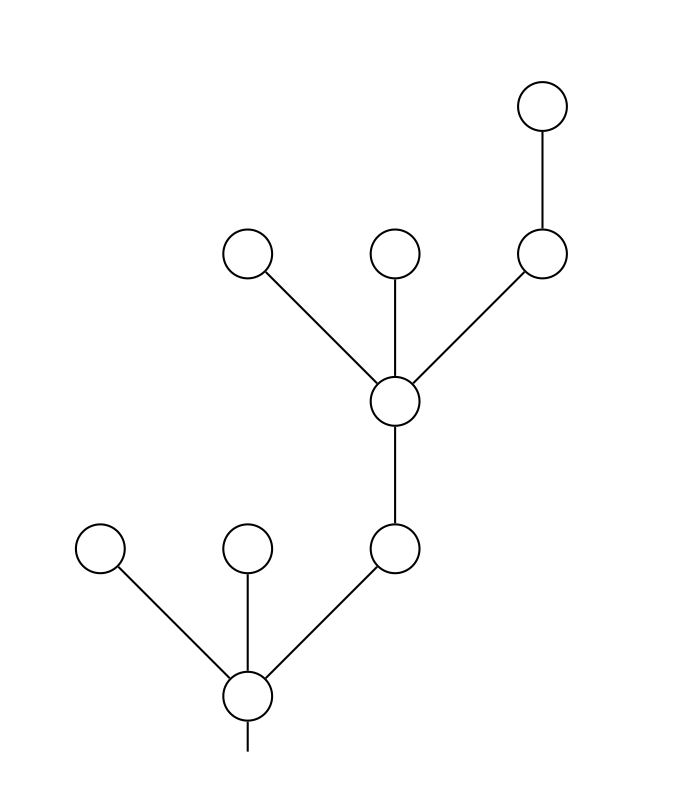}
\caption{Example structure learned by network for 10 tasks.}
\label{structure}
\end{center}
\end{figure}

\noindent Multiple nodes could be attached to an existing node or backbone. The network grows a tree-like structure (Fig.~\ref{structure}), with each path from the root to the leaf containing the parameters utilized for the specific task. We select the most similar task according to a similarity measure $s$ that will be discussed later. We attach a new node to the backbone if all tasks are dissimilar. The general rule for network expansion could be written as:

\begin{equation}
    \mathcal{F}^t = 
    \begin{cases}
    \mathcal{F}^t_C \circ \mathcal{F}^t_S \circ \mathcal{F}^{t-i}_{B-S}  &  if \exists_{i<t} s(i) \geq \bar{s} \\
    \mathcal{F}^t_C \circ \mathcal{F}^t_S \circ \mathcal{F}_B(.)  &  if \forall_{i < t} s(i) < \bar{s}
    \end{cases}
\end{equation}

\noindent where $\bar{s}$ is the predefined similarity threshold, and $\mathcal{F}^{t-i}_{B-S}$ are layers for task $t-i$ with task-specific layers removed.

\subsection{Task similarity} \label{task_similarity}

A crucial element of the proposed method is establishing similarity between two tasks. It is not trivial as we cannot access datasets used for training previous tasks. We may only use data for new tasks $\mathcal{D}_t$ and model trained for previous tasks $\mathcal{F}^{t-1}$. For this reason, we select the most similar task based on the entropy of predictions. The entropy for the previous task $i$ could be calculated as:

\begin{equation}
    H_i = - \frac{1}{|\mathcal{D}_i|} \sum_{x \in \mathcal{D}_i} \sum_{1 \leq k \leq |\mathcal{Y}_i|} \left(\mathcal{F}^{t-1}(x)_k \right)log \left(\mathcal{F}^{t-1}(x)_k \right)
\end{equation}

\noindent where $|\mathcal{Y}_i|$ is number of classes for task $i$ and $\mathcal{F}^{t-1}(x)_k$ is $k$-th output of network $\mathcal{F}^{t-1}$.
The maximum entropy value may vary depending on the number of classes for the $i$th task, but the number of classes for each task could be different. Therefore, we cannot directly compare entropy for different paths in our network. To unify, we compute the maximum value of entropy for a given task:

\begin{equation}
    % H_{i}^{max} = - \sum_{j = 0}^{|\mathcal{Y}_i|} \frac{1}{|\mathcal{Y}_i|} log \frac{1}{|\mathcal{Y}_i|}
    H_{i}^{max} = log |\mathcal{Y}_i|
\end{equation}

\noindent To compare the task similarity of each previous task $i$ to the new one, we use following quantity:
\begin{equation}
    s(i) = 1 - \frac{H_i}{H_{i}^{max}}
\end{equation}

\noindent the entropy ratio has values in $[0, 1]$ interval, regardless of the number of classes for each task, therefore proposed similarity score is also in the same interval. 

\subsection{Time complexity} \label{time_complexity}

The proposed algorithm for tree growing increases the time complexity of the proposed method, as for each new task, we must compute entropy for previous nodes. It could be optimized by utilizing tree structure and storing intermediate activations that could be later passed to multiple nodes. Another alternative is to use a multi-GPU system, as entropy calculation could be paralleled to multiple processes. During training, the time complexity of forward pass could be written as $O(\mathcal{F}_B + d \mathcal{F}_S + \mathcal{F}_C) $, where $O(\mathcal{F}_B)$ is the time complexity of backbone training, $d$ is the depth of the network path for the current task, $O(\mathcal{F}_S)$ is the complexity of the shared part and $O(\mathcal{F}_T)$ is the complexity of the task-specific part. It means that crucial for time complexity is the depth of the current path $d$. If a tree is wider, $d$ is smaller. The width of a tree could be controlled to some extent by changing hyperparameter $\bar{s}$. With a higher threshold $\bar{s}$, the model will more often attach new nodes directly to the backbone, increasing the width instead of the tree's depth. The time complexity of the backward pass depends only on the current node, as all previous weights are frozen and do not require gradient computation. For this reason, the backward pass is independent of the current depth $d$.
\subsection{Very deep neural networks} \label{very_deep_neural_networks}

Model depth cannot be increased indefinitely. For implementation reasons, very deep models could not be fitted into single GPU memory. The first problem could be solved by using model parallelism \cite{NIPS2012_6aca9700} and assigning different parts of networks to different GPUs. Here we may simply solve this problem by introducing the maximum depth limit. When the network grows to maximum depth, the latest nodes are not considered when selecting where new layers should be attached.

Another problem caused by the indefinite increasing in network depth is the eventual occurrence of overfitting. We avoid this problem by employing the stochastic depth \cite{DBLP:journals/corr/HuangSLSW16}. Our model is based on ResNet with stochastic depth blocks in the form of:

\begin{equation}
    \mathcal{R}^l(x) = x + p_l \cdot \mathcal{C}(x)
\end{equation}

\noindent where $\mathcal{R}^{l}$ is residual block at depth $l$, $p_l$ is a Bernoulli random variable, and $\mathcal{C}$ is the set of convolution layers, nonlinear activations and batch normalization applied to input $x$. In principle, each residual block could be randomly skipped during network forward propagation. For this reason, stochastic depth is a form of normalization useful for training deeper models \cite{DBLP:journals/corr/abs-1905-11946}. The frequency of using each block could be determined based on two parameters. First is the probability of using the first layer $p_1$. The first layer contains important information about general features that the next layers utilize to create more complex data representations. For this reason, random exclusion of this layer could introduce too much noise and, consequently, impede the network training process. To avoid these consequences, $p_1$ usually takes high values such as 1. The second parameter is the probability $p_L$ of utilizing the last block $L$. This layer produces more abstract representations and could be skipped more easily. For this reason, $p_L$ could take lower values. Parameters for Bernoulli random variables of all in-between blocks could be computed by linearly scaling between $p_1$ and $p_L$:

\begin{equation}
\label{eqn:p_update}
    p_l = p_1 - l \cdot \frac{p_1 - p_L}{L-1}
\end{equation}

\noindent Adding new nodes to our network increases the number of residual blocks. In this case, we are using Eq.~\ref{eqn:p_update} to update the Bernoulli random variables for each residual block in the current path. This form of regularisation should, in principle, allow for training networks with even thousands of layers.

\subsection{Batch normalization} \label{batch_normalization}

Batch normalization \cite{DBLP:journals/corr/IoffeS15} is used in many modern neural networks \cite{DBLP:journals/corr/HeZRS15,DBLP:journals/corr/abs-1905-11946,DBLP:journals/corr/XieGDTH16,DBLP:journals/corr/SzegedyIV16,DBLP:journals/corr/abs-2104-00298}. This layer computes activation statistics heavily dependent on data distribution and contains learnable parameters interacting with activation statistics. In a life-long learning scenario, data distribution varies for each task. In the preliminary experiment, we noticed that this could impact our model behavior in the long term. Also, the performance of neural networks with batch normalization is dependent on batch size \cite{DBLP:journals/corr/abs-1803-08494}. With increased depth, a smaller batch size could be required to fit the model into available GPU memory. For this reason, we swap batch normalization layers in our model to instance normalization \cite{DBLP:journals/corr/UlyanovVL16}.

\subsection{Model inference} \label{model_inference}

At inference time, we utilize information about task index $t$ to select the proper path through the already built network structure and utilize all layers to produce probability estimates for each class. The final classification is based on selecting a class with a higher probability:

\begin{equation}
     \hat{y_i} = \mathop{argmax}\limits_{1 \leq j \leq |\mathcal{Y_i}| } p(y_j | x_i)
\end{equation}

\section{Experimental evaluation}
This part describes details of experimental research, including their goals, experimental setup, used baselines, base architecture, the hyperparameter tuning process, and the obtained results. We will summarize this section with an in-depth discussion of the results.

\subsection{Research questions}
%We state that the following research questions (RQs) are to be answered as a result of conducting the experimental research:
The experiments aim to answer the following research questions:
\begin{enumerate}
    \item Could a gradual increase of network depth be used for continual learning?
    \item What is the benefit of utilizing pre-trained architectures in life-long learning?
    \item Does the proposed method of building a tree structure provides improvement over stacking new nodes on top of each other?
    \item Do significant shifts in task datasets distributions make learning with increasing depth more difficult?
\end{enumerate}

\subsection{Experimental protocol} \label{experimental_protocol}

We conduct broad experiments with multiple datasets and commonly used baselines in continual learning. To provide fair evaluation, we provide results for all methods with and without pretraining.
Additionally, we performed ablation studies to examine what parts of our solution are important.

The proposed method was implemented in Python programming language with Pytorch \cite{NEURIPS2019_9015} and Avalanche \cite{lomonaco2021avalanche} libraries. The source code is available online \footnote{https://github.com/w4k2/increasing-depth-life-long} to allow the reproducibility of our research.

\subsection{Datasets} \label{datasets}

We chose commonly used benchmarks for continual learning:
\begin{itemize}
    \item Split CIFAR \cite{Krizhevsky09learningmultiple} created by splitting CIFAR100 dataset \cite{cifar} into 20 different tasks, 5 classes for each task. 
    \item Split Tiny ImageNet dataset \cite{Le2015TinyIV} with 20 tasks and 5 classes for each task.
    \item CORe50 \cite{DBLP:journals/corr/LomonacoM17} is dataset and benchmark created specifically for continual learning. In our experiment, we use CORe50 dataset and split it into 10 tasks with 5 classes each.
\end{itemize}.

To evaluate performance in environments where the shift in data distribution between different tasks is bigger we utilize benchmarks with different computer vision datasets as separate task.
For this purpose, we employ 5-dataset \cite{DBLP:journals/corr/abs-2003-09553} consisting of commonly used datasets, namely: SVHN \cite{svhn}, CIFAR10 \cite{Krizhevsky09learningmultiple}, MNIST \cite{LeCun2005TheMD}, FashionMNIST \cite{DBLP:journals/corr/abs-1708-07747}, and NOTMNIST \cite{nmnist}.

All images in experiments were resized to 64x64 pixels for models with the ResNet base model. For CIFAR, Tiny ImageNet, and CORe50 datasets standard set of augmentations was used i.e., random resized crop and random horizontal flip for training data. For SVHN, MNSIT, FashionMNIST, and NOTMNIST we only resize image and apply data normalization.

\subsection{Baselines} \label{baselines}

The proposed method is based on architecture expansion and should be primarily compared to other existing methods in this group. However, to thoroughly evaluate the proposed method, we choose several commonly used methods that represent each group of continual learning methods introduced in related works:
\begin{itemize}
    \item An upper bound of possible performance with ResNet model. We use models trained with cumulative datasets, encompassing data from all tasks seen so far.
    \item Experience Replay (ER) \cite{DBLP:journals/corr/abs-1902-10486} - a method that creates buffers with samples from previous tasks. 
    \item A-GEM \cite{DBLP:journals/corr/abs-1812-00420} - a more advanced version of replay, created for online continual learning. It utilizes gradient projections to limit performance degradation on previous tasks. 
    \item PNN \cite{DBLP:journals/corr/RusuRDSKKPH16} is the most similar method to ours, expanding the overall network structure with a new backbone for each task. 
    \item EWC \cite{DBLP:journals/corr/KirkpatrickPRVD16} - the regularization-based method. 
    \item LwF \cite{DBLP:journals/corr/LiH16e} - a method that uses pseudo labels generated by classification heads for previous tasks.
    \item HAT \cite{DBLP:journals/corr/abs-1801-01423} - a method that learns attention masks for weights and prevents updates for weights already used in previous tasks.
    \item CAT \cite{10.5555/3495724.3497277} - a method based on HAT that utilizes the shared parameters in the form of a knowledge base modified based on learnable task embeddings and dot-product attention for using learned weights for previous tasks. 
\end{itemize}

\subsection{Architecture} \label{architecture}

We employed ResNet18 \cite{DBLP:journals/corr/HeZRS15} as the base model for all methods, except for methods that utilize custom network structures like PNN or HAT. 
To provide an in-depth study of the impact of pre-trained weights, we train all methods that utilize ResNet18 with and without ImageNet pretraining. All algorithms using custom network structures, namely PNN, HAT, and CAT, are trained only from scratch with randomly initialized weights. 
In the case of PNN, the network architecture is different from the standard computer vision backbone. For this reason, we only swap convolution used by the authors of the original paper for residual blocks used in ResNet18 and match the network depth.
We trained replay, A-GEM, PNN, baseline methods, and our method with Adam \cite{adam} optimizer, and LwF, EWC with SGD \cite{ruder2016overview}.

\subsection{Hyperparameters} \label{hyperparameters}

We performed grid search hyperparameter optimization with the following parameters: learning rate, weight decay, and the number of epochs. Other algorithm-specific parameters were found manually before the grid search. We used the strength of regularization in EWC equal to 1000; we allowed memory size for 250 patterns per task for A-GEM and ER; distillation and temperature equal to 1 for LwF, and $p_1=1$ and $p_L=0.5$ for our method. According to \cite{DBLP:journals/corr/abs-1812-00420}, and \cite{DBLP:journals/corr/abs-1906-02425}, we allowed hyperparameter search only for the first three tasks and reported final results for all tasks. We trained networks for multiple learning epochs, as the number of learning examples in each task was usually too small to train for only one epoch.

\subsection{Threats to validity}

%Our paper lacks statistical analysis with proper hypothesis testing. This issue was, to some degree, mitigated by reporting metrics values averaged over several independent runs with standard deviation. In most cases, the difference in performance between different algorithms is greater than the standard deviation of obtained results. Therefore analysis of the results should be valid.
One of the possible threats is the excess of techniques introduced in the paper, which may lead to an over-complicated solution. We have performed the ablation study to verify the effectiveness and necessity of all novel techniques introduced in this paper. 
Conclusions from research papers could also be false due to the wrong implementation of experimental code or baseline method. We have minimized this risk by using the Avalanche library with the implementation of several baselines. In the case of baselines not implemented within Avalanche, we have used the code provided by the authors of the original papers. 
Another possible threat to validity is the hyperparameter tuning process. It was shown previously that the hyperparameter tuning process could severely impact the outcomes of the machine learning experiment \cite{DBLP:journals/corr/abs-2003-08505,DBLP:journals/corr/MelisDB17,DBLP:journals/corr/abs-1907-06902}. For this reason we have performed the careful hyperparameter tuning process with extensive gird search optimization for learning rate, weight decay and number of epochs. The first two hyperparameters are important for the training process of neural networks in general, while we found the third one to be particularly important for continual learning, as it could provide a tradeoff between forgetting and plasticity. We have consulted the original paper or implementation to select the best values for hyperparameters specific to a given algorithm. In principle, a more extensive hyperparameter search could impact the obtained results. However, we had only a limited computational budget at our disposal.

% When it comes to external validity, 
We have performed experiments with a standard set of benchmarks used in continual learning. One may wonder whether these benchmarks represent the real data distribution shifts that can occur worldwide. CIFAR100 and TinyImageNet are older datasets created for general-purpose image recognition and later reused as continual benchmarks. This kind of dataset "recycling" was criticized in \cite{2021_3b8a6142}, as datasets meant for one task should not be repurposed for another. The CORe50 dataset is a newer dataset created from the beginning with continual learning in mind. Experiments on this dataset could increase the validity of our results. Please also note that we have omitted the usage of Permuted MNIST benchmark \cite{DBLP:journals/corr/KirkpatrickPRVD16} from our experiments because this dataset was criticized previously \cite{farquhar2019robust} due to unrealistic changes in data distribution between multiple tasks. All the datasets mentioned so far were created with data from a single source that was later arbitrarily split into several tasks. The real-life distribution shifts could be more drastic in scope. For example, changing the data acquisition process could severely impact the data. Therefore, we have included in our experiments a 5-datasets benchmark to evaluate the performance of proposed algorithms in scenarios where data shifts are more drastic. 
When applying continual learning to existing systems, one should also consider that in real life, the switching point from one distribution to another is unknown. Moreover, it does not have to be an abrupt change. It could also be a gradual process with an interleaving period when data from both tasks is available to the learner. Such a scenario is not well-studied in continual learning literature, and only few papers have considered this setup.

\subsection{Results}

This section presents the results of conducted experiments according to the protocol described in the previous section.

\subsubsection{Results on standard benchmarks}

We report average task accuracy after training (Acc) and forgetting measure (FM) \cite{DBLP:journals/corr/abs-1801-10112}. All metrics were averaged over five runs. For all datasets, the upper bound sets limit performance for convolutional models. Results of our experiments are presented in Tab.~\ref{tab:results_cifar},\ref{tab:results_tinyimagenet},\ref{tab:results_core}.

% \begin{table*}[t]
%     \caption{Results for selected dataset.}
%     \hspace{-1.5in}
%     % \centering
%     % \begin{center}
%         \input{tables/main_results}
%     \label{tab:results_all}
%     % \end{center}
% \end{table*}

\begin{table}[t]
    \caption{Results for CIFAR100 dataset.}
    \begin{center}
        \begin{tabular}{c|cc}
            \hline
            \multicolumn{3}{c}{\textbf{w/o pretraining}} \\
            \hline
            \textbf{Method} & \textbf{Acc} & \textbf{FM} \\
            \hline
        Upperbound                                       & 0.8467±0.0143 & 0.0183±0.005  \\
        EWC \cite{DBLP:journals/corr/KirkpatrickPRVD16}  & 0.5963±0.016  & 0.12±0.0166   \\
        ER \cite{DBLP:journals/corr/abs-1902-10486}      & 0.7522±0.0219 & 0.0884±0.0142 \\
        A-GEM \cite{DBLP:journals/corr/abs-1812-00420}   & 0.503±0.0495  & 0.3297±0.054  \\
        PNN \cite{DBLP:journals/corr/RusuRDSKKPH16}     & \textbf{0.7838±0.0137} & \textbf{0.0±0.0} \\
        LWF \cite{DBLP:journals/corr/LiH16e}             & 0.3666±0.0459 & 0.4619±0.0429 \\
        HAT \cite{DBLP:journals/corr/abs-1801-01423}    & 0.748±0.0137  & 0.0001±0.0001 \\
        CAT \cite{10.5555/3495724.3497277} & 0.6994±0.0072 & 0.0801±0.0151 \\
        Ours                                            & 0.73±0.0137   & \textbf{0.0±0.0} \\

        \hline
        \multicolumn{3}{c}{\textbf{w/ pretraining}} \\
        \hline
        \textbf{Method} & \textbf{Acc} & \textbf{FM} \\
        \hline
        Upperbound                                       & 0.8702±0.0125 & 0.0137±0.0018 \\
        EWC \cite{DBLP:journals/corr/KirkpatrickPRVD16}  & 0.6543±0.0226 & 0.1921±0.0205 \\
        ER \cite{DBLP:journals/corr/abs-1902-10486}      & \textbf{0.8152±0.0146} & 0.0555±0.0079 \\
        A-GEM \cite{DBLP:journals/corr/abs-1812-00420}   & 0.5227±0.0471 & 0.3411±0.0456 \\
        LWF \cite{DBLP:journals/corr/LiH16e}             & 0.5379±0.0374 & 0.2992±0.025  \\
        Ours                                            & 0.8091±0.0102 & \textbf{0.0±0.0} \\
            \hline
        \end{tabular}
    \label{tab:results_cifar}
    \end{center}
\end{table}

\begin{table}[t]
    \caption{Results for Tiny ImageNet dataset.}
    \begin{center}
        \begin{tabular}{c|cc}
            \hline
            \multicolumn{3}{c}{\textbf{w/o pretraining}} \\
            \hline
            \textbf{Method} & \textbf{Acc} & \textbf{FM} \\
            \hline
        Upperbound                                      &  0.5289±0.0212 & 0.0831±0.0165  \\
        EWC \cite{DBLP:journals/corr/KirkpatrickPRVD16} &  0.4324±0.0111 & 0.0318±0.0029   \\
        ER \cite{DBLP:journals/corr/abs-1902-10486}     &  0.5219±0.0075 & 0.1265±0.0097 \\
        A-GEM \cite{DBLP:journals/corr/abs-1812-00420}  &  0.349±0.0617  & 0.3036±0.0609  \\
        PNN \cite{DBLP:journals/corr/RusuRDSKKPH16}     & \textbf{0.5577±0.0054} & \textbf{0.0±0.0} \\
        LWF \cite{DBLP:journals/corr/LiH16e}            &  0.2844±0.0194 & 0.352±0.0145 \\
        HAT \cite{DBLP:journals/corr/abs-1801-01423}    & 0.53±0.0071 & 0.0001±0.0001 \\
        CAT \cite{10.5555/3495724.3497277} & 0.4567±0.0051 & 0.0597±0.0147 \\
        Ours                                            & 0.5218±0.0074 & \textbf{0.0±0.0} \\

        \hline
        \multicolumn{3}{c}{\textbf{w/ pretraining}} \\
        \hline
        \textbf{Method} & \textbf{Acc} & \textbf{FM} \\
        \hline
        Upperbound                                      &  0.7±0.0034    & 0.012±0.0021 \\
        EWC \cite{DBLP:journals/corr/KirkpatrickPRVD16} &  0.5399±0.017  & 0.0899±0.0079 \\
        ER \cite{DBLP:journals/corr/abs-1902-10486}     &  0.5735±0.0143 & 0.1197±0.0114 \\
        A-GEM \cite{DBLP:journals/corr/abs-1812-00420}  &  0.4208±0.0174 & 0.2699±0.0202 \\
        LWF \cite{DBLP:journals/corr/LiH16e}            &  0.3983±0.0076 & 0.3141±0.0065  \\
        Ours                                            & \textbf{0.5811±0.005}  & \textbf{0.0±0.0} \\
            \hline
        \end{tabular}
    \label{tab:results_tinyimagenet}
    \end{center}
\end{table}

\begin{table}[t]
    \caption{Results for CORe50 dataset.}
    \begin{center}
        \begin{tabular}{c|cc}
            \hline
            \multicolumn{3}{c}{\textbf{w/o pretraining}} \\
            \hline
            \textbf{Method} & \textbf{Acc} & \textbf{FM} \\
            \hline
        Upperbound                                      & 0.9957±0.0025 & 0.0031±0.0025  \\
        EWC \cite{DBLP:journals/corr/KirkpatrickPRVD16} & 0.6325±0.0469 & 0.3663±0.0475   \\
        ER \cite{DBLP:journals/corr/abs-1902-10486}     & 0.8848±0.0238 & 0.1143±0.024 \\
        A-GEM \cite{DBLP:journals/corr/abs-1812-00420}  & 0.6567±0.0339 & 0.339±0.033  \\
        PNN \cite{DBLP:journals/corr/RusuRDSKKPH16}     & \textbf{0.9986±0.0005} & \textbf{0.0±0.0} \\
        LWF \cite{DBLP:journals/corr/LiH16e}            & 0.5035±0.0385 & 0.4963±0.0386 \\
        HAT \cite{DBLP:journals/corr/abs-1801-01423}    & 0.8393±0.3197 & 0.0001±0.0001 \\
        CAT \cite{10.5555/3495724.3497277} & 0.9932±0.0012  & 0.003±0.001 \\
        Ours                                            & 0.9955±0.002           & \textbf{0.0±0.0} \\

        \hline
        \multicolumn{3}{c}{\textbf{w/ pretraining}} \\
        \hline
        \textbf{Method} & \textbf{Acc} & \textbf{FM} \\
        \hline
        Upperbound                                      & 0.996±0.0033  & 0.0034±0.0033 \\
        EWC \cite{DBLP:journals/corr/KirkpatrickPRVD16} & 0.8082±0.0556 & 0.171±0.0655 \\
        ER \cite{DBLP:journals/corr/abs-1902-10486}     & 0.895±0.022   & 0.1004±0.0213 \\
        A-GEM \cite{DBLP:journals/corr/abs-1812-00420}  & 0.6383±0.0406 & 0.3289±0.0427 \\
        LWF \cite{DBLP:journals/corr/LiH16e}            & 0.5571±0.0562 & 0.4412±0.0561  \\
        Ours                                            & \textbf{0.9955±0.002}  & \textbf{0.0±0.0} \\
            \hline
        \end{tabular}
    \label{tab:results_core}
    \end{center}
\end{table}

We split our results into two parts to provide a clear comparison between results obtained for methods with and without pretraining. Progressive Neural Networks obtained the best performance on all benchmarks without pretraining. Because standard convolutional blocks used in the original PNN architecture were swapped for residual blocks, this indicates that methods based on expanding the networks could be considered strong baselines with high performance and low forgetting. 
Experience Replay, Hard Attention to Task, and the proposed method obtain similar accuracy. This suggests that our method of pretraining is on par with the most commonly used methods in the literature.

When the use of pre-trained weights is allowed, we note improvement in accuracy for all methods and at the same time, no significant difference in the Forgetting Measure for most algorithms. When comparing the accuracy obtained by different methods, we observe that Experience Replay and the proposed method may obtain close results that other baselines cannot match with or without pretraining. This suggests that the proposed method is a valid alternative to replay-based methods in situations where data from previous tasks cannot be stored due to, for example, privacy concerns. For CORe50, many methods obtain results close to the Upperbound. In this case, the proposed method obtains performance on par with the best algorithms.

\begin{figure}[]
\centering
\hspace*{-1.0in}
\includegraphics[width=18cm]{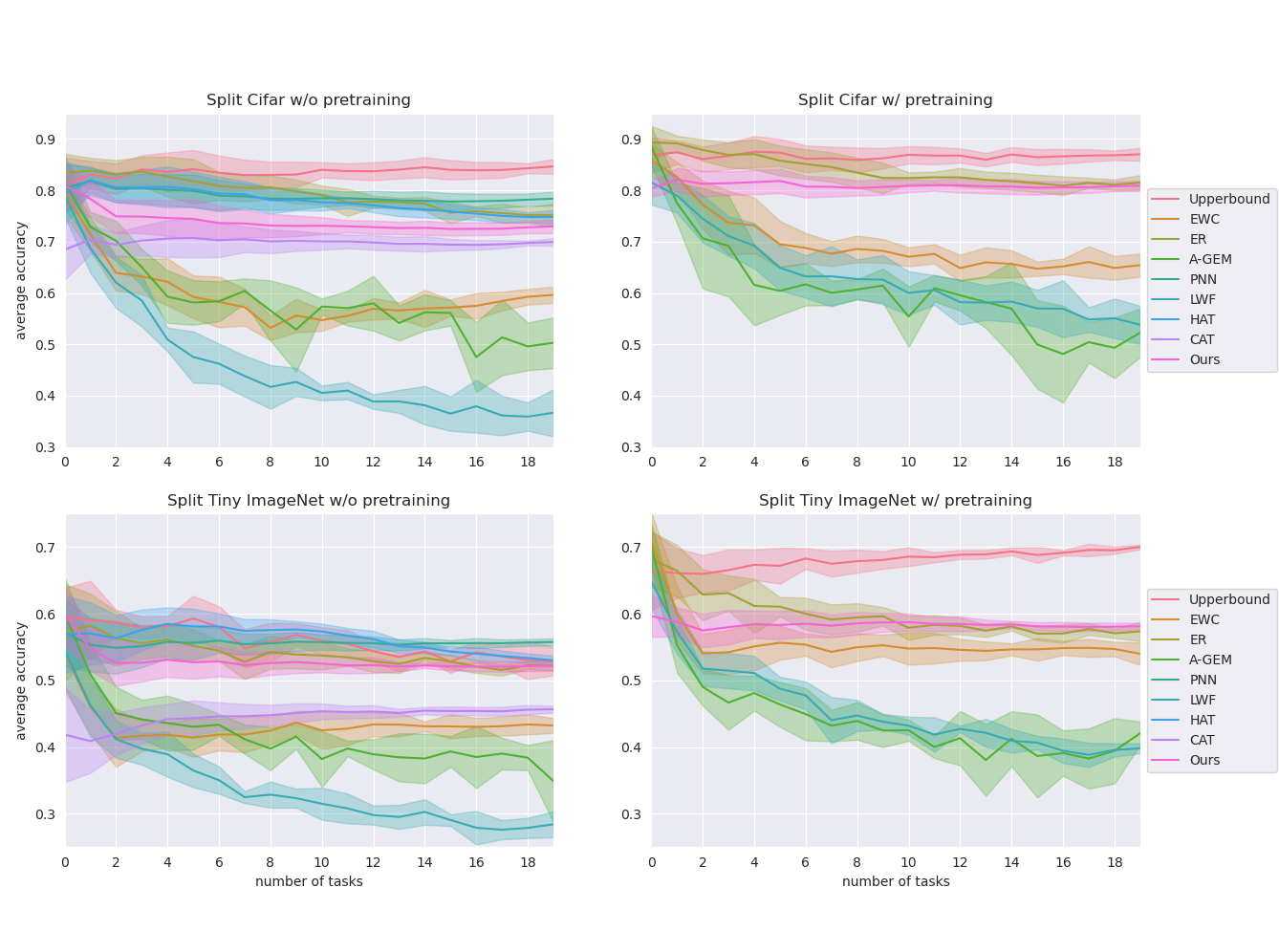}
\caption{Average accuracy for Split Cifar and Split Tiny Imagenet datasets. Results are averaged over 5 runs. Standard deviation is marked by shaded area for all methods.}
\label{fig:res}
\end{figure}

More detailed analysis may be carried out when we consider the dynamics of the learning process across multiple tasks. For this purpose, in Fig.~\ref{fig:res}, we plot average accuracy with standard deviation for all tasks from Split Cifar and Split Tiny ImageNet datasets.
Our method obtains stable accuracy that does not degrade over time. Also, the standard deviation is low compared to other methods. The same observations could be made for PNN. However, due to the custom structure of the network, it cannot benefit from ImageNet pretraining. 
Experience Replay at begin of training obtains comparable accuracy to the Upperbound. However, over time, the performance becomes worse. One may expect that with more tasks, ER would degrade even further.
Regularization-based approaches suffer from high forgetting during the training process. ImageNet pretraining allows for higher accuracy. However, the overall trend is the same - performance quickly degrades over time.

To compare algorithms in Tab.~\ref{tab:parameters}, we specify the number of trainable parameters after training with 20 tasks on the Split Cifar benchmark. Methods based on regularisation or rehearsal utilize a standard ResNet18 network with 11M parameters. All algorithms with custom architectures have larger number of parameters after training, with the proposed method having the highest count. The overall high count of parameters may be attributed to the number of weights in each block. In the ResNet18 network 75\% of all weights are located in the last residual block. We are using the last two blocks for new nodes in our architecture.
This may be alarming, however, there are a few important properties that we need to have in mind when comparing a number of parameters. Firstly our method has a high count of parameters for all tasks, but most of these parameters are not stored in GPU memory during training. To train our network, we need only a path from the tree's root to the leaf that contains the currently used parameters. Secondly, the weights for previous tasks are frozen. For this reason, they do not require complete gradient computation for the entire network. In the backward pass, we compute gradients only for the newest node. For this reason, both times of training and memory requirements are not severely affected by network expansion. 
In contrast, for the PNN network, each task creates its network. Networks for previously currently used tasks must be loaded into GPU memory. This prohibits the utilization of modern neural network architectures, as during training, multiple networks must be present in graphics card memory.

\begin{table}[t]
    \centering
        \caption{Number of paramters for each method after training with 20 tasks on Split CIFAR.}
        \begin{tabular}{c|c}
            \hline
            \textbf{Method} & \textbf{\#parameters} \\
            \hline
        Upperbound                                      &           11M \\
        EWC \cite{DBLP:journals/corr/KirkpatrickPRVD16} &           11M \\
        ER \cite{DBLP:journals/corr/abs-1902-10486}     &           11M \\
        A-GEM \cite{DBLP:journals/corr/abs-1812-00420}  &           11M \\
        PNN \cite{DBLP:journals/corr/RusuRDSKKPH16}     &           39M \\
        LWF \cite{DBLP:journals/corr/LiH16e}            &           11M \\
        HAT \cite{DBLP:journals/corr/abs-1801-01423}    &           28M \\
        CAT \cite{10.5555/3495724.3497277} & 90M \\
        Ours                                            &           193M \\
        \hline
        \end{tabular}
    \label{tab:parameters}
\end{table}

\subsubsection{Ablation studies}

We performed ablation studies on the Split Cifar benchmark to estimate what parts of our solution contribute most to obtained results. 
First, we have removed stochastic depth from residual blocks in our network. Next, we removed the pretraining. We also study the impact of the proposed network growing strategy based on entropy similarity. Instead of selecting the most similar tasks, we attach new layers to the latest node in a network, creating a sequential structure. The results are presented in Tab.~\ref{tab:ablation}.

We observe a 1.16\% drop in accuracy when swapping stochastic depth for conventional residual blocks. This proves that stochastic depth is a useful form of regularisation in our method that allows for training with increasing layers.
The 7.99\% drop in accuracy after removing ImageNet pretraining is probably due to the low amount of training data available. Each task in the Split Cifar benchmark has only 2500 training samples, which is insufficient to train general representations in the convolutional neural network.
When selecting where a new node should be attached, entropy-based similarity task selection improves the performance by approximately 1.7\% when compared to the naive stacking of new layers.

Analyzing the results of ablations studies, one may observe that the most impact on the results has an ImageNet pretraining. This observation is in line with results already published in literature \cite{DBLP:journals/corr/abs-2112-09153}. Continual learning algorithms could benefit from introducing pre-trained weights. The same finding was present in the previous part of our experiments. An increase in accuracy was noted for all methods after utilizing pretraining. 
Our method freezes pre-trained weights, allowing for better use of pretraining without the risk of forgetting. This is probably why the performance difference between ER and our method gets smaller as we add ImageNet pretraining.

\newcommand{\cmark}{\ding{51}}
\newcommand{\xmark}{\ding{55}}

\begin{table}
\caption{Ablation study results.}
\begin{center}
    \begin{tabular}{ccc|c}
    \hline
    \textbf{stochastic depth} & \textbf{pretraining} & \textbf{entropy similarity} & \textbf{accuracy} \\
    \hline
    \cmark & \cmark & \cmark & 0.815 \\ 
    \xmark & \cmark & \cmark & 0.799 \\
    \xmark & \xmark & \cmark & 0.72  \\
    \xmark & \xmark & \xmark & 0.703 \\
    \hline
    \end{tabular}
\label{tab:ablation}
\end{center}
\end{table}

\subsection{Multi-dataset benchmarks}

We also tested our method in a more demanding environment where each task was a separate dataset encompassing a different domain. We want to evaluate whether bigger differences in task data distributions impact learning with the proposed method. We selected three methods from previous experiments: Elastic Weight Consolidation, Experience Replay, and A-GEM. We have chosen these methods based on their popularity in literature.
Results are presented in Tab.~\ref{tab:results_multi}. Replay and proposed method benefit from using pre-trained weights, with average accuracy, increased by approximately 2\%. We do not observe an improvement in accuracy for EWC and A-GEM after applying pre-trained weights. The proposed method performs well in this setting, with over 10\% gain in accuracy over experience replay, and still provides no forgetting.

% 500 rehersal samples per task

\begin{table}
    \caption{Results for 5-dataset benchmark.}
    \begin{center}
        \begin{tabular}{c|cc|cc}
        \hline
         \multirow{2}{*}{\textbf{Method}} & \multicolumn{2}{c}{\textbf{w/o pretraining}} & \multicolumn{2}{c}{\textbf{w/ pretraining}} \\
         \cline{2-5}
         & \textbf{Acc} & \textbf{FM} & \textbf{Acc} & \textbf{FM} \\
        \hline
        %  PNN      & 0.6548 & 0.3492 & -         & -         \\
        %  EWC &
        %  ER       & 0.7614  & 0.2048 & 0.7753 & 0.215 \\
        %  ours     & \textbf{0.8776} & \textbf{0}      & \textbf{0.8934} & \textbf{0}     \\
         EWC      & 0.6774 & 0.3491 & 0.678 & 0.2284 \\
         ER       & 0.7614 & 0.2048 & 0.7753 & 0.215 \\
         A-GEM    & 0.6292 & 0.3977 & 0.6015 & 0.4455 \\
         Ours     & \textbf{0.8776} & \textbf{0}        & \textbf{0.8934} & \textbf{0}        \\

        \hline
        \end{tabular}
    \label{tab:results_multi}
    \end{center}
\end{table}

\subsection{Lessons learned}

The first research question concerned the possibility of increasing depth for continual learning.
The results obtained in our work prove that increasing neural network depth could be beneficial for continual learning when correctly utilized. As shown with ablation studies, the naive stacking of new layers does not provide much gain. This is in line with results from Iman et al. \cite{mirzadeh2022architecture}, where the impact of architecture on continual learning performance was studied. Authors suggested that using wider, not deeper, networks in a life-long learning setup is more beneficial. 

The experiments show that the proposed method obtains competitive performance, especially when using pre-trained weights. When discussing results obtained in our work, it is essential to note that some baselines received worse results than those presented in the literature. This is probably due to two factors. First is the utilization of larger networks in our experiments. Some methods \cite{DBLP:journals/corr/abs-1812-00420} were developed with a Reduced Resnet18 - network with the same structure as ResNet18 but three times fewer channels or with smaller convolutional networks. Changing neural network architecture could degrade performance. The second reason is that we allow longer training, up to 50 epochs, during hyperparameters tuning. Even if we consider better results from the literature, the proposed method still performs well (RQ1 answered).

The second research question was regarding the impact of pretraining for continual learning. 
From our experiments, we may deduce that using pretrained weights improves performance. This is in line with the previous result from \cite{DBLP:journals/corr/abs-2112-09153}; however, there is a significant difference between the performance gap in previous work and our experiments. In \cite{DBLP:journals/corr/abs-2112-09153}, pretraining can improve accuracy from 0.1 to 0.2 for computer vision benchmarks. In our experiments, the difference in accuracy obtained with training with and without pretraining for most methods is below 0.1. We may make the same observation about multi-dataset setup - a gain in accuracy was also observed, but it was not of the magnitude reported in the previous paper. 
A study on the impact of pretraining from \cite{DBLP:journals/corr/abs-2112-09153} utilized only three algorithms. Furthermore, the same learning rate was used for all methods, along with the smaller size of the rehearsal buffer, a lower regularization rate used for Elastic Weight Consolidation, and a different image size of 224x224. This difference in hyperparameters and experimental setup could lead to a lower performance for methods without pretraining. Our results also suggest a significant benefit from employing pretraining weights; however, the full extent of this improvement should be more thoroughly evaluated. This calls for further research on the impact of pretraining on continual learning performance, emphasizing hyperparameter influence (RQ2 answered).

The next question stated in our research was regarding the proposed method for building the tree structure of our model and its comparison to alternatives. 
Ablation studies suggest that naive sequential stacking of new layers does not provide good results. Utilization of task similarity obtains better performance. It confirms the intuition behind our method. During sequential training on new data, activations learned for the previous tasks may be irrelevant to the current task. In this situation, there is a need for a mechanism that allows for selecting the more useful representations for a new node (RQ3 answered). 

The last research question was about learning in live-long learning environments with more significant shifts between subsequent tasks.
When there is a substantial difference between task distributions, one could expect our method to fail. Especially when no pre-trained weights are used, the second and subsequent tasks must use common backbone activations learned for the first task. This could hinder performance when there is a substantial difference in data distribution between tasks. Surprisingly, as shown by our experiments with the 5-datasets benchmark, that is not the case. Suppose representations obtained during training with the first task are general enough. They could be "useful" for new nodes, even if there is a substantial difference in the distributions of subsequent tasks (RQ4 answered).

\section{Practicality considerations}

Continual learning could be used in many domains where data distribution is evolving, and deployed models should be capable to adapt to changes in the environment. The area that could benefit greatly from the introduction of continual learning is automatic medical diagnosis. Due to the low robustness of current deep learning to changes in data distribution \cite{doi:10.34133/2022/9807590}, experts advise gradual extending of model capabilities by careful evaluation of algorithm performance for new equipment, hospitals, and data formats \cite{Liu2022}. Here continual learning could help prevent the retraining of models multiple times, which could be costly and time-consuming. The same argument could be used when it comes to the adaptation of large language models to new tasks. Continual learning could allow for domain adaptation with a lower loss of general capabilities of the original model.
Another possible application area is the development of autonomous mobile robots \cite{Chen:2018}. An autonomous agent will inevitably discover some objects or situations not well-represented in training data. Mobile robots should be capable to adapt to newly encountered data while not losing previous abilities. 
In principle, the perfect application of continual learning would include a single deployment of a model on an edge device and continuous adaptation based on user-specific data \cite{DBLP:journals/corr/abs-2105-13127}. Such an approach could limit the bandwidth for sending user data and model predictions, lower the number of compute needed to obtain predictions on servers, and reduce privacy concerns. Unfortunately, current continual learning algorithms are far from the efficiency required for such solution.

Due to its nature, continual learning is well suited for applications where no additional effort for labeling is needed. Such areas may include but are not limited to recommender systems, dynamic pricing, estimating the time of arrival, or stock price prediction. In applications without easy access to new labels, continual learning could be used in combination with test-time adaptation \cite{wang2022continual} or self-supervised learning \cite{tang2023practical}.

The implementation of continual learning from overall system perspective is described in Machine Learning Engineering textbook \cite{huyen2022designing}. First, the trained model is stored and used to predict incoming samples. We will refer to this version as a champion model. Next, a replica of the champion model will be continually updated with incoming data. This is the challenger model that ought to replace the champion model. After a predefined period, we evaluate the performance of both models to decide whether the challenger model should replace the champion. Evaluation could be done offline with a predetermined test set or with production testing methods such as A/B testing or canary deployment \cite{Kohavi2017TheSP}. Periodically new model could be retrained from scratch to include more newly collected data.
This whole approach is also known as Stateful Training \cite{semola2022continuallearningasaservice} because the same model is continually updated with incoming data. The opposite method is Stateless Retraining, where a new model is retrained from scratch in a fixed period, with no consideration for accumulated knowledge from the previously trained model.

The proposed method could be deployed on a server-side. Deployment on edge devices could be difficult due to memory requirements. However, most modern continual learning methods are unsuitable for edge deployment for the same reason. We have tested that on a single-GPU computer with 24Gb of GPU memory, our method could handle more than 100 tasks. 24Gbs is not the high end of memory size for modern GPUs, therefore our model should be able to model more tasks with better hardware. Also, on the multi-GPU servers, we may use techniques, such as model parallelism, that allow splitting models computation across different graphic cards.
Also, please note that our method reduces the need for storing two copies of the same model in the Stateful Training framework. We may create only the new node that is trained, while previously trained weights remain frozen, so there is no need to copy the same model. 
For this reason, the proposed method is well-suited for applications where the model must be adapted to new data, however, due to memory constraints, we would not recommend using it with the fastest data streams with model update intervals in the range of a couple of minutes.

\section{Conclusion}

We proposed a novel method for continual learning based on the increasing depth of neural networks. We indicate where new layers should be attached by computing the entropy of each head prediction for a new task. We created a tree-like structure that could grow based on task similarity based on this information. Our method is similar in principle to PNN; however, PNN allocates a lot of memory for the whole network structure during the learning process. The proposed method alleviates this by adding only part of a network for a new task and utilizing a subset of previously trained weights. At the same time, we may retain the benefit of PNN, such as no forgetting guaranteed by design, without needing a memory buffer.

Conducted experiments suggest that the introduced algorithm obtained good results, especially for datasets with distribution close to real images. In a more challenging setup with a single computer vision dataset as a separate task, our method outperforms Experience Replay.  

Future work may include adding the ability to select tasks in the class incremental scenario, where no label of incoming task is provided. Also, methods for tree structure pruning should be developed to limit memory usage.

\section*{Acknowledgment}
%It will be added if the paper is accepted.
This work is supported by the CEUS-UNISONO programme, which has received funding from the National Science Centre, Poland under grant agreement No. 2020/02/Y/ST6/00037.

\bibliographystyle{elsarticle-num-names} 
\bibliography{conf}

\begin{thebibliography}{66}
\expandafter\ifx\csname natexlab\endcsname\relax\def\natexlab#1{#1}\fi
\providecommand{\url}[1]{\texttt{#1}}
\providecommand{\href}[2]{#2}
\providecommand{\path}[1]{#1}
\providecommand{\DOIprefix}{doi:}
\providecommand{\ArXivprefix}{arXiv:}
\providecommand{\URLprefix}{URL: }
\providecommand{\Pubmedprefix}{pmid:}
\providecommand{\doi}[1]{\href{http://dx.doi.org/#1}{\path{#1}}}
\providecommand{\Pubmed}[1]{\href{pmid:#1}{\path{#1}}}
\providecommand{\bibinfo}[2]{#2}
\ifx\xfnm\relax \def\xfnm[#1]{\unskip,\space#1}\fi
%Type = Article
\bibitem[{He et~al.(2015)He, Zhang, Ren, and Sun}]{DBLP:journals/corr/HeZRS15}
\bibinfo{author}{K.~He}, \bibinfo{author}{X.~Zhang}, \bibinfo{author}{S.~Ren},
  \bibinfo{author}{J.~Sun},
\newblock \bibinfo{title}{Deep residual learning for image recognition},
\newblock \bibinfo{journal}{CoRR} \bibinfo{volume}{abs/1512.03385}
  (\bibinfo{year}{2015}). \URLprefix \url{http://arxiv.org/abs/1512.03385}.
  \href{http://arxiv.org/abs/1512.03385}{{\tt arXiv:1512.03385}}.
%Type = Article
\bibitem[{Huang et~al.(2016)Huang, Sun, Liu, Sedra, and
  Weinberger}]{DBLP:journals/corr/HuangSLSW16}
\bibinfo{author}{G.~Huang}, \bibinfo{author}{Y.~Sun}, \bibinfo{author}{Z.~Liu},
  \bibinfo{author}{D.~Sedra}, \bibinfo{author}{K.~Q. Weinberger},
\newblock \bibinfo{title}{Deep networks with stochastic depth},
\newblock \bibinfo{journal}{CoRR} \bibinfo{volume}{abs/1603.09382}
  (\bibinfo{year}{2016}). \URLprefix \url{http://arxiv.org/abs/1603.09382}.
  \href{http://arxiv.org/abs/1603.09382}{{\tt arXiv:1603.09382}}.
%Type = Article
\bibitem[{Tan and Le(2019)}]{DBLP:journals/corr/abs-1905-11946}
\bibinfo{author}{M.~Tan}, \bibinfo{author}{Q.~V. Le},
\newblock \bibinfo{title}{Efficientnet: Rethinking model scaling for
  convolutional neural networks},
\newblock \bibinfo{journal}{CoRR} \bibinfo{volume}{abs/1905.11946}
  (\bibinfo{year}{2019}). \URLprefix \url{http://arxiv.org/abs/1905.11946}.
  \href{http://arxiv.org/abs/1905.11946}{{\tt arXiv:1905.11946}}.
%Type = Article
\bibitem[{Russakovsky et~al.(2014)Russakovsky, Deng, Su, Krause, Satheesh, Ma,
  Huang, Karpathy, Khosla, Bernstein, Berg, and
  Fei{-}Fei}]{DBLP:journals/corr/RussakovskyDSKSMHKKBBF14}
\bibinfo{author}{O.~Russakovsky}, \bibinfo{author}{J.~Deng},
  \bibinfo{author}{H.~Su}, \bibinfo{author}{J.~Krause},
  \bibinfo{author}{S.~Satheesh}, \bibinfo{author}{S.~Ma},
  \bibinfo{author}{Z.~Huang}, \bibinfo{author}{A.~Karpathy},
  \bibinfo{author}{A.~Khosla}, \bibinfo{author}{M.~S. Bernstein},
  \bibinfo{author}{A.~C. Berg}, \bibinfo{author}{L.~Fei{-}Fei},
\newblock \bibinfo{title}{Imagenet large scale visual recognition challenge},
\newblock \bibinfo{journal}{CoRR} \bibinfo{volume}{abs/1409.0575}
  (\bibinfo{year}{2014}). \URLprefix \url{http://arxiv.org/abs/1409.0575}.
  \href{http://arxiv.org/abs/1409.0575}{{\tt arXiv:1409.0575}}.
%Type = Article
\bibitem[{Xie et~al.(2016)Xie, Girshick, Doll{\'{a}}r, Tu, and
  He}]{DBLP:journals/corr/XieGDTH16}
\bibinfo{author}{S.~Xie}, \bibinfo{author}{R.~B. Girshick},
  \bibinfo{author}{P.~Doll{\'{a}}r}, \bibinfo{author}{Z.~Tu},
  \bibinfo{author}{K.~He},
\newblock \bibinfo{title}{Aggregated residual transformations for deep neural
  networks},
\newblock \bibinfo{journal}{CoRR} \bibinfo{volume}{abs/1611.05431}
  (\bibinfo{year}{2016}). \URLprefix \url{http://arxiv.org/abs/1611.05431}.
  \href{http://arxiv.org/abs/1611.05431}{{\tt arXiv:1611.05431}}.
%Type = Article
\bibitem[{Szegedy et~al.(2016)Szegedy, Ioffe, and
  Vanhoucke}]{DBLP:journals/corr/SzegedyIV16}
\bibinfo{author}{C.~Szegedy}, \bibinfo{author}{S.~Ioffe},
  \bibinfo{author}{V.~Vanhoucke},
\newblock \bibinfo{title}{Inception-v4, inception-resnet and the impact of
  residual connections on learning},
\newblock \bibinfo{journal}{CoRR} \bibinfo{volume}{abs/1602.07261}
  (\bibinfo{year}{2016}). \URLprefix \url{http://arxiv.org/abs/1602.07261}.
  \href{http://arxiv.org/abs/1602.07261}{{\tt arXiv:1602.07261}}.
%Type = Article
\bibitem[{Zeiler and Fergus(2013)}]{DBLP:journals/corr/ZeilerF13}
\bibinfo{author}{M.~D. Zeiler}, \bibinfo{author}{R.~Fergus},
\newblock \bibinfo{title}{Visualizing and understanding convolutional
  networks},
\newblock \bibinfo{journal}{CoRR} \bibinfo{volume}{abs/1311.2901}
  (\bibinfo{year}{2013}). \URLprefix \url{http://arxiv.org/abs/1311.2901}.
  \href{http://arxiv.org/abs/1311.2901}{{\tt arXiv:1311.2901}}.
%Type = Article
\bibitem[{Bozinovski(2020)}]{Bozinovski2020ReminderOT}
\bibinfo{author}{S.~Bozinovski},
\newblock \bibinfo{title}{Reminder of the first paper on transfer learning in
  neural networks, 1976},
\newblock \bibinfo{journal}{Informatica (Slovenia)} \bibinfo{volume}{44}
  (\bibinfo{year}{2020}).
%Type = Article
\bibitem[{Kolesnikov et~al.(2019)Kolesnikov, Beyer, Zhai, Puigcerver, Yung,
  Gelly, and Houlsby}]{DBLP:journals/corr/abs-1912-11370}
\bibinfo{author}{A.~Kolesnikov}, \bibinfo{author}{L.~Beyer},
  \bibinfo{author}{X.~Zhai}, \bibinfo{author}{J.~Puigcerver},
  \bibinfo{author}{J.~Yung}, \bibinfo{author}{S.~Gelly},
  \bibinfo{author}{N.~Houlsby},
\newblock \bibinfo{title}{Large scale learning of general visual
  representations for transfer},
\newblock \bibinfo{journal}{CoRR} \bibinfo{volume}{abs/1912.11370}
  (\bibinfo{year}{2019}). \URLprefix \url{http://arxiv.org/abs/1912.11370}.
  \href{http://arxiv.org/abs/1912.11370}{{\tt arXiv:1912.11370}}.
%Type = Article
\bibitem[{Li and Hoiem(2016)}]{DBLP:journals/corr/LiH16e}
\bibinfo{author}{Z.~Li}, \bibinfo{author}{D.~Hoiem},
\newblock \bibinfo{title}{Learning without forgetting},
\newblock \bibinfo{journal}{CoRR} \bibinfo{volume}{abs/1606.09282}
  (\bibinfo{year}{2016}). \URLprefix \url{http://arxiv.org/abs/1606.09282}.
  \href{http://arxiv.org/abs/1606.09282}{{\tt arXiv:1606.09282}}.
%Type = Article
\bibitem[{French(1999)}]{catastrophic_forgetting}
\bibinfo{author}{R.~French},
\newblock \bibinfo{title}{Catastrophic forgetting in connectionist networks},
\newblock \bibinfo{journal}{Trends in cognitive sciences} \bibinfo{volume}{3}
  (\bibinfo{year}{1999}) \bibinfo{pages}{128--135}.
  \DOIprefix\doi{10.1016/S1364-6613(99)01294-2}.
%Type = Book
\bibitem[{Chen et~al.(2018)Chen, Liu, Brachman, Stone, and Rossi}]{Chen:2018}
\bibinfo{author}{Z.~Chen}, \bibinfo{author}{B.~Liu},
  \bibinfo{author}{R.~Brachman}, \bibinfo{author}{P.~Stone},
  \bibinfo{author}{F.~Rossi}, \bibinfo{title}{Lifelong Machine Learning},
  \bibinfo{edition}{2nd} ed., \bibinfo{publisher}{Morgan {\&} Claypool
  Publishers}, \bibinfo{year}{2018}.
%Type = Article
\bibitem[{Ditzler et~al.(2015)Ditzler, Roveri, Alippi, and Polikar}]{7296710}
\bibinfo{author}{G.~Ditzler}, \bibinfo{author}{M.~Roveri},
  \bibinfo{author}{C.~Alippi}, \bibinfo{author}{R.~Polikar},
\newblock \bibinfo{title}{Learning in nonstationary environments: A survey},
\newblock \bibinfo{journal}{IEEE Computational Intelligence Magazine}
  \bibinfo{volume}{10} (\bibinfo{year}{2015}) \bibinfo{pages}{12--25}.
  \DOIprefix\doi{10.1109/MCI.2015.2471196}.
%Type = Article
\bibitem[{Raghavan and Balaprakash(2021)}]{DBLP:journals/corr/abs-2109-14035}
\bibinfo{author}{K.~Raghavan}, \bibinfo{author}{P.~Balaprakash},
\newblock \bibinfo{title}{Formalizing the generalization-forgetting trade-off
  in continual learning},
\newblock \bibinfo{journal}{CoRR} \bibinfo{volume}{abs/2109.14035}
  (\bibinfo{year}{2021}). \URLprefix \url{https://arxiv.org/abs/2109.14035}.
  \href{http://arxiv.org/abs/2109.14035}{{\tt arXiv:2109.14035}}.
%Type = Misc
\bibitem[{Bremner et~al.(2013)Bremner, Lewkowicz, and Spence}]{human_learning}
\bibinfo{author}{A.~Bremner}, \bibinfo{author}{D.~Lewkowicz},
  \bibinfo{author}{C.~Spence}, \bibinfo{title}{Multisensory development},
  \bibinfo{year}{2013}.
  \DOIprefix\doi{10.1093/acprof:oso/9780199586059.003.0001}.
%Type = Article
\bibitem[{Parisi et~al.(2018)Parisi, Kemker, Part, Kanan, and
  Wermter}]{DBLP:journals/corr/abs-1802-07569}
\bibinfo{author}{G.~I. Parisi}, \bibinfo{author}{R.~Kemker},
  \bibinfo{author}{J.~L. Part}, \bibinfo{author}{C.~Kanan},
  \bibinfo{author}{S.~Wermter},
\newblock \bibinfo{title}{Continual lifelong learning with neural networks: {A}
  review},
\newblock \bibinfo{journal}{CoRR} \bibinfo{volume}{abs/1802.07569}
  (\bibinfo{year}{2018}). \URLprefix \url{http://arxiv.org/abs/1802.07569}.
  \href{http://arxiv.org/abs/1802.07569}{{\tt arXiv:1802.07569}}.
%Type = Article
\bibitem[{Simonyan and Zisserman(2014)}]{vgg}
\bibinfo{author}{K.~Simonyan}, \bibinfo{author}{A.~Zisserman},
\newblock \bibinfo{title}{Very deep convolutional networks for large-scale
  image recognition},
\newblock \bibinfo{journal}{arXiv 1409.1556}  (\bibinfo{year}{2014}).
%Type = Article
\bibitem[{Szegedy et~al.(2015)Szegedy, Vanhoucke, Ioffe, Shlens, and
  Wojna}]{DBLP:journals/corr/SzegedyVISW15}
\bibinfo{author}{C.~Szegedy}, \bibinfo{author}{V.~Vanhoucke},
  \bibinfo{author}{S.~Ioffe}, \bibinfo{author}{J.~Shlens},
  \bibinfo{author}{Z.~Wojna},
\newblock \bibinfo{title}{Rethinking the inception architecture for computer
  vision},
\newblock \bibinfo{journal}{CoRR} \bibinfo{volume}{abs/1512.00567}
  (\bibinfo{year}{2015}). \URLprefix \url{http://arxiv.org/abs/1512.00567}.
  \href{http://arxiv.org/abs/1512.00567}{{\tt arXiv:1512.00567}}.
%Type = Article
\bibitem[{Tan and Le(2021)}]{DBLP:journals/corr/abs-2104-00298}
\bibinfo{author}{M.~Tan}, \bibinfo{author}{Q.~V. Le},
\newblock \bibinfo{title}{Efficientnetv2: Smaller models and faster training},
\newblock \bibinfo{journal}{CoRR} \bibinfo{volume}{abs/2104.00298}
  (\bibinfo{year}{2021}). \URLprefix \url{https://arxiv.org/abs/2104.00298}.
  \href{http://arxiv.org/abs/2104.00298}{{\tt arXiv:2104.00298}}.
%Type = Article
\bibitem[{Girshick et~al.(2013)Girshick, Donahue, Darrell, and
  Malik}]{DBLP:journals/corr/GirshickDDM13}
\bibinfo{author}{R.~B. Girshick}, \bibinfo{author}{J.~Donahue},
  \bibinfo{author}{T.~Darrell}, \bibinfo{author}{J.~Malik},
\newblock \bibinfo{title}{Rich feature hierarchies for accurate object
  detection and semantic segmentation},
\newblock \bibinfo{journal}{CoRR} \bibinfo{volume}{abs/1311.2524}
  (\bibinfo{year}{2013}). \URLprefix \url{http://arxiv.org/abs/1311.2524}.
  \href{http://arxiv.org/abs/1311.2524}{{\tt arXiv:1311.2524}}.
%Type = Article
\bibitem[{He et~al.(2017)He, Gkioxari, Doll{\'{a}}r, and
  Girshick}]{DBLP:journals/corr/HeGDG17}
\bibinfo{author}{K.~He}, \bibinfo{author}{G.~Gkioxari},
  \bibinfo{author}{P.~Doll{\'{a}}r}, \bibinfo{author}{R.~B. Girshick},
\newblock \bibinfo{title}{Mask {R-CNN}},
\newblock \bibinfo{journal}{CoRR} \bibinfo{volume}{abs/1703.06870}
  (\bibinfo{year}{2017}). \URLprefix \url{http://arxiv.org/abs/1703.06870}.
  \href{http://arxiv.org/abs/1703.06870}{{\tt arXiv:1703.06870}}.
%Type = Article
\bibitem[{Zoph and Le(2016)}]{DBLP:journals/corr/ZophL16}
\bibinfo{author}{B.~Zoph}, \bibinfo{author}{Q.~V. Le},
\newblock \bibinfo{title}{Neural architecture search with reinforcement
  learning},
\newblock \bibinfo{journal}{CoRR} \bibinfo{volume}{abs/1611.01578}
  (\bibinfo{year}{2016}). \URLprefix \url{http://arxiv.org/abs/1611.01578}.
  \href{http://arxiv.org/abs/1611.01578}{{\tt arXiv:1611.01578}}.
%Type = Article
\bibitem[{Zoph et~al.(2017)Zoph, Vasudevan, Shlens, and
  Le}]{DBLP:journals/corr/ZophVSL17}
\bibinfo{author}{B.~Zoph}, \bibinfo{author}{V.~Vasudevan},
  \bibinfo{author}{J.~Shlens}, \bibinfo{author}{Q.~V. Le},
\newblock \bibinfo{title}{Learning transferable architectures for scalable
  image recognition},
\newblock \bibinfo{journal}{CoRR} \bibinfo{volume}{abs/1707.07012}
  (\bibinfo{year}{2017}). \URLprefix \url{http://arxiv.org/abs/1707.07012}.
  \href{http://arxiv.org/abs/1707.07012}{{\tt arXiv:1707.07012}}.
%Type = Article
\bibitem[{Kirkpatrick et~al.(2016)Kirkpatrick, Pascanu, Rabinowitz, Veness,
  Desjardins, Rusu, Milan, Quan, Ramalho, Grabska{-}Barwinska, Hassabis,
  Clopath, Kumaran, and Hadsell}]{DBLP:journals/corr/KirkpatrickPRVD16}
\bibinfo{author}{J.~Kirkpatrick}, \bibinfo{author}{R.~Pascanu},
  \bibinfo{author}{N.~C. Rabinowitz}, \bibinfo{author}{J.~Veness},
  \bibinfo{author}{G.~Desjardins}, \bibinfo{author}{A.~A. Rusu},
  \bibinfo{author}{K.~Milan}, \bibinfo{author}{J.~Quan},
  \bibinfo{author}{T.~Ramalho}, \bibinfo{author}{A.~Grabska{-}Barwinska},
  \bibinfo{author}{D.~Hassabis}, \bibinfo{author}{C.~Clopath},
  \bibinfo{author}{D.~Kumaran}, \bibinfo{author}{R.~Hadsell},
\newblock \bibinfo{title}{Overcoming catastrophic forgetting in neural
  networks},
\newblock \bibinfo{journal}{CoRR} \bibinfo{volume}{abs/1612.00796}
  (\bibinfo{year}{2016}). \URLprefix \url{http://arxiv.org/abs/1612.00796}.
  \href{http://arxiv.org/abs/1612.00796}{{\tt arXiv:1612.00796}}.
%Type = Article
\bibitem[{Rusu et~al.(2016)Rusu, Rabinowitz, Desjardins, Soyer, Kirkpatrick,
  Kavukcuoglu, Pascanu, and Hadsell}]{DBLP:journals/corr/RusuRDSKKPH16}
\bibinfo{author}{A.~A. Rusu}, \bibinfo{author}{N.~C. Rabinowitz},
  \bibinfo{author}{G.~Desjardins}, \bibinfo{author}{H.~Soyer},
  \bibinfo{author}{J.~Kirkpatrick}, \bibinfo{author}{K.~Kavukcuoglu},
  \bibinfo{author}{R.~Pascanu}, \bibinfo{author}{R.~Hadsell},
\newblock \bibinfo{title}{Progressive neural networks},
\newblock \bibinfo{journal}{CoRR} \bibinfo{volume}{abs/1606.04671}
  (\bibinfo{year}{2016}). \URLprefix \url{http://arxiv.org/abs/1606.04671}.
  \href{http://arxiv.org/abs/1606.04671}{{\tt arXiv:1606.04671}}.
%Type = Article
\bibitem[{Lee et~al.(2017)Lee, Yoon, Yang, and
  Hwang}]{DBLP:journals/corr/abs-1708-01547}
\bibinfo{author}{J.~Lee}, \bibinfo{author}{J.~Yoon}, \bibinfo{author}{E.~Yang},
  \bibinfo{author}{S.~J. Hwang},
\newblock \bibinfo{title}{Lifelong learning with dynamically expandable
  networks},
\newblock \bibinfo{journal}{CoRR} \bibinfo{volume}{abs/1708.01547}
  (\bibinfo{year}{2017}). \URLprefix \url{http://arxiv.org/abs/1708.01547}.
  \href{http://arxiv.org/abs/1708.01547}{{\tt arXiv:1708.01547}}.
%Type = Article
\bibitem[{Li et~al.(2019)Li, Zhou, Wu, Socher, and
  Xiong}]{DBLP:journals/corr/abs-1904-00310}
\bibinfo{author}{X.~Li}, \bibinfo{author}{Y.~Zhou}, \bibinfo{author}{T.~Wu},
  \bibinfo{author}{R.~Socher}, \bibinfo{author}{C.~Xiong},
\newblock \bibinfo{title}{Learn to grow: {A} continual structure learning
  framework for overcoming catastrophic forgetting},
\newblock \bibinfo{journal}{CoRR} \bibinfo{volume}{abs/1904.00310}
  (\bibinfo{year}{2019}). \URLprefix \url{http://arxiv.org/abs/1904.00310}.
  \href{http://arxiv.org/abs/1904.00310}{{\tt arXiv:1904.00310}}.
%Type = Article
\bibitem[{Chaudhry et~al.(2019)Chaudhry, Rohrbach, Elhoseiny, Ajanthan,
  Dokania, Torr, and Ranzato}]{DBLP:journals/corr/abs-1902-10486}
\bibinfo{author}{A.~Chaudhry}, \bibinfo{author}{M.~Rohrbach},
  \bibinfo{author}{M.~Elhoseiny}, \bibinfo{author}{T.~Ajanthan},
  \bibinfo{author}{P.~K. Dokania}, \bibinfo{author}{P.~H.~S. Torr},
  \bibinfo{author}{M.~Ranzato},
\newblock \bibinfo{title}{Continual learning with tiny episodic memories},
\newblock \bibinfo{journal}{CoRR} \bibinfo{volume}{abs/1902.10486}
  (\bibinfo{year}{2019}). \URLprefix \url{http://arxiv.org/abs/1902.10486}.
  \href{http://arxiv.org/abs/1902.10486}{{\tt arXiv:1902.10486}}.
%Type = Article
\bibitem[{Lopez{-}Paz and Ranzato(2017)}]{DBLP:journals/corr/Lopez-PazR17}
\bibinfo{author}{D.~Lopez{-}Paz}, \bibinfo{author}{M.~Ranzato},
\newblock \bibinfo{title}{Gradient episodic memory for continuum learning},
\newblock \bibinfo{journal}{CoRR} \bibinfo{volume}{abs/1706.08840}
  (\bibinfo{year}{2017}). \URLprefix \url{http://arxiv.org/abs/1706.08840}.
  \href{http://arxiv.org/abs/1706.08840}{{\tt arXiv:1706.08840}}.
%Type = Article
\bibitem[{Chaudhry et~al.(2018)Chaudhry, Ranzato, Rohrbach, and
  Elhoseiny}]{DBLP:journals/corr/abs-1812-00420}
\bibinfo{author}{A.~Chaudhry}, \bibinfo{author}{M.~Ranzato},
  \bibinfo{author}{M.~Rohrbach}, \bibinfo{author}{M.~Elhoseiny},
\newblock \bibinfo{title}{Efficient lifelong learning with {A-GEM}},
\newblock \bibinfo{journal}{CoRR} \bibinfo{volume}{abs/1812.00420}
  (\bibinfo{year}{2018}). \URLprefix \url{http://arxiv.org/abs/1812.00420}.
  \href{http://arxiv.org/abs/1812.00420}{{\tt arXiv:1812.00420}}.
%Type = Inproceedings
\bibitem[{Dean et~al.(2012)Dean, Corrado, Monga, Chen, Devin, Mao, Ranzato,
  Senior, Tucker, Yang, Le, and Ng}]{NIPS2012_6aca9700}
\bibinfo{author}{J.~Dean}, \bibinfo{author}{G.~Corrado},
  \bibinfo{author}{R.~Monga}, \bibinfo{author}{K.~Chen},
  \bibinfo{author}{M.~Devin}, \bibinfo{author}{M.~Mao}, \bibinfo{author}{M.~a.
  Ranzato}, \bibinfo{author}{A.~Senior}, \bibinfo{author}{P.~Tucker},
  \bibinfo{author}{K.~Yang}, \bibinfo{author}{Q.~Le}, \bibinfo{author}{A.~Ng},
\newblock \bibinfo{title}{Large scale distributed deep networks},
\newblock in: \bibinfo{editor}{F.~Pereira}, \bibinfo{editor}{C.~J.~C. Burges},
  \bibinfo{editor}{L.~Bottou}, \bibinfo{editor}{K.~Q. Weinberger} (Eds.),
  \bibinfo{booktitle}{Advances in Neural Information Processing Systems},
  volume~\bibinfo{volume}{25}, \bibinfo{publisher}{Curran Associates, Inc.},
  \bibinfo{year}{2012}.
%Type = Article
\bibitem[{Ioffe and Szegedy(2015)}]{DBLP:journals/corr/IoffeS15}
\bibinfo{author}{S.~Ioffe}, \bibinfo{author}{C.~Szegedy},
\newblock \bibinfo{title}{Batch normalization: Accelerating deep network
  training by reducing internal covariate shift},
\newblock \bibinfo{journal}{CoRR} \bibinfo{volume}{abs/1502.03167}
  (\bibinfo{year}{2015}). \URLprefix \url{http://arxiv.org/abs/1502.03167}.
  \href{http://arxiv.org/abs/1502.03167}{{\tt arXiv:1502.03167}}.
%Type = Article
\bibitem[{Wu and He(2018)}]{DBLP:journals/corr/abs-1803-08494}
\bibinfo{author}{Y.~Wu}, \bibinfo{author}{K.~He},
\newblock \bibinfo{title}{Group normalization},
\newblock \bibinfo{journal}{CoRR} \bibinfo{volume}{abs/1803.08494}
  (\bibinfo{year}{2018}). \URLprefix \url{http://arxiv.org/abs/1803.08494}.
  \href{http://arxiv.org/abs/1803.08494}{{\tt arXiv:1803.08494}}.
%Type = Article
\bibitem[{Ulyanov et~al.(2016)Ulyanov, Vedaldi, and
  Lempitsky}]{DBLP:journals/corr/UlyanovVL16}
\bibinfo{author}{D.~Ulyanov}, \bibinfo{author}{A.~Vedaldi},
  \bibinfo{author}{V.~S. Lempitsky},
\newblock \bibinfo{title}{Instance normalization: The missing ingredient for
  fast stylization},
\newblock \bibinfo{journal}{CoRR} \bibinfo{volume}{abs/1607.08022}
  (\bibinfo{year}{2016}). \URLprefix \url{http://arxiv.org/abs/1607.08022}.
  \href{http://arxiv.org/abs/1607.08022}{{\tt arXiv:1607.08022}}.
%Type = Incollection
\bibitem[{Paszke et~al.(2019)Paszke, Gross, Massa, Lerer, Bradbury, Chanan,
  Killeen, Lin, Gimelshein, Antiga, Desmaison, Kopf, Yang, DeVito, Raison,
  Tejani, Chilamkurthy, Steiner, Fang, Bai, and Chintala}]{NEURIPS2019_9015}
\bibinfo{author}{A.~Paszke}, \bibinfo{author}{S.~Gross},
  \bibinfo{author}{F.~Massa}, \bibinfo{author}{A.~Lerer},
  \bibinfo{author}{J.~Bradbury}, \bibinfo{author}{G.~Chanan},
  \bibinfo{author}{T.~Killeen}, \bibinfo{author}{Z.~Lin},
  \bibinfo{author}{N.~Gimelshein}, \bibinfo{author}{L.~Antiga},
  \bibinfo{author}{A.~Desmaison}, \bibinfo{author}{A.~Kopf},
  \bibinfo{author}{E.~Yang}, \bibinfo{author}{Z.~DeVito},
  \bibinfo{author}{M.~Raison}, \bibinfo{author}{A.~Tejani},
  \bibinfo{author}{S.~Chilamkurthy}, \bibinfo{author}{B.~Steiner},
  \bibinfo{author}{L.~Fang}, \bibinfo{author}{J.~Bai},
  \bibinfo{author}{S.~Chintala},
\newblock \bibinfo{title}{Pytorch: An imperative style, high-performance deep
  learning library},
\newblock in: \bibinfo{editor}{H.~Wallach}, \bibinfo{editor}{H.~Larochelle},
  \bibinfo{editor}{A.~Beygelzimer}, \bibinfo{editor}{F.~d\textquotesingle
  Alch\'{e}-Buc}, \bibinfo{editor}{E.~Fox}, \bibinfo{editor}{R.~Garnett}
  (Eds.), \bibinfo{booktitle}{Advances in Neural Information Processing Systems
  32}, \bibinfo{publisher}{Curran Associates, Inc.}, \bibinfo{year}{2019}, pp.
  \bibinfo{pages}{8024--8035}.
%Type = Inproceedings
\bibitem[{Lomonaco et~al.(2021)Lomonaco, Pellegrini, Cossu, Carta, Graffieti,
  Hayes, Lange, Masana, Pomponi, van~de Ven, Mundt, She, Cooper, Forest,
  Belouadah, Calderara, Parisi, Cuzzolin, Tolias, Scardapane, Antiga, Amhad,
  Popescu, Kanan, van~de Weijer, Tuytelaars, Bacciu, and
  Maltoni}]{lomonaco2021avalanche}
\bibinfo{author}{V.~Lomonaco}, \bibinfo{author}{L.~Pellegrini},
  \bibinfo{author}{A.~Cossu}, \bibinfo{author}{A.~Carta},
  \bibinfo{author}{G.~Graffieti}, \bibinfo{author}{T.~L. Hayes},
  \bibinfo{author}{M.~D. Lange}, \bibinfo{author}{M.~Masana},
  \bibinfo{author}{J.~Pomponi}, \bibinfo{author}{G.~van~de Ven},
  \bibinfo{author}{M.~Mundt}, \bibinfo{author}{Q.~She},
  \bibinfo{author}{K.~Cooper}, \bibinfo{author}{J.~Forest},
  \bibinfo{author}{E.~Belouadah}, \bibinfo{author}{S.~Calderara},
  \bibinfo{author}{G.~I. Parisi}, \bibinfo{author}{F.~Cuzzolin},
  \bibinfo{author}{A.~Tolias}, \bibinfo{author}{S.~Scardapane},
  \bibinfo{author}{L.~Antiga}, \bibinfo{author}{S.~Amhad},
  \bibinfo{author}{A.~Popescu}, \bibinfo{author}{C.~Kanan},
  \bibinfo{author}{J.~van~de Weijer}, \bibinfo{author}{T.~Tuytelaars},
  \bibinfo{author}{D.~Bacciu}, \bibinfo{author}{D.~Maltoni},
\newblock \bibinfo{title}{Avalanche: an end-to-end library for continual
  learning},
\newblock in: \bibinfo{booktitle}{Proceedings of IEEE Conference on Computer
  Vision and Pattern Recognition}, 2nd Continual Learning in Computer Vision
  Workshop, \bibinfo{year}{2021}.
%Type = Techreport
\bibitem[{Krizhevsky(2009)}]{Krizhevsky09learningmultiple}
\bibinfo{author}{A.~Krizhevsky}, \bibinfo{title}{Learning multiple layers of
  features from tiny images}, \bibinfo{type}{Technical Report},
  \bibinfo{year}{2009}.
%Type = Article
\bibitem[{Krizhevsky(2012)}]{cifar}
\bibinfo{author}{A.~Krizhevsky},
\newblock \bibinfo{title}{Learning multiple layers of features from tiny
  images},
\newblock \bibinfo{journal}{University of Toronto}  (\bibinfo{year}{2012}).
%Type = Inproceedings
\bibitem[{Le and Yang(2015)}]{Le2015TinyIV}
\bibinfo{author}{Y.~Le}, \bibinfo{author}{X.~S. Yang},
\newblock \bibinfo{title}{Tiny imagenet visual recognition challenge},
\newblock \bibinfo{year}{2015}.
%Type = Article
\bibitem[{Lomonaco and Maltoni(2017)}]{DBLP:journals/corr/LomonacoM17}
\bibinfo{author}{V.~Lomonaco}, \bibinfo{author}{D.~Maltoni},
\newblock \bibinfo{title}{Core50: a new dataset and benchmark for continuous
  object recognition},
\newblock \bibinfo{journal}{CoRR} \bibinfo{volume}{abs/1705.03550}
  (\bibinfo{year}{2017}). \URLprefix \url{http://arxiv.org/abs/1705.03550}.
  \href{http://arxiv.org/abs/1705.03550}{{\tt arXiv:1705.03550}}.
%Type = Article
\bibitem[{Ebrahimi et~al.(2020)Ebrahimi, Meier, Calandra, Darrell, and
  Rohrbach}]{DBLP:journals/corr/abs-2003-09553}
\bibinfo{author}{S.~Ebrahimi}, \bibinfo{author}{F.~Meier},
  \bibinfo{author}{R.~Calandra}, \bibinfo{author}{T.~Darrell},
  \bibinfo{author}{M.~Rohrbach},
\newblock \bibinfo{title}{Adversarial continual learning},
\newblock \bibinfo{journal}{CoRR} \bibinfo{volume}{abs/2003.09553}
  (\bibinfo{year}{2020}). \URLprefix \url{https://arxiv.org/abs/2003.09553}.
  \href{http://arxiv.org/abs/2003.09553}{{\tt arXiv:2003.09553}}.
%Type = Inproceedings
\bibitem[{Netzer et~al.(2011)Netzer, Wang, Coates, Bissacco, Wu, and Ng}]{svhn}
\bibinfo{author}{Y.~Netzer}, \bibinfo{author}{T.~Wang},
  \bibinfo{author}{A.~Coates}, \bibinfo{author}{A.~Bissacco},
  \bibinfo{author}{B.~Wu}, \bibinfo{author}{A.~Y. Ng},
\newblock \bibinfo{title}{Reading digits in natural images with unsupervised
  feature learning},
\newblock in: \bibinfo{booktitle}{NIPS Workshop on Deep Learning and
  Unsupervised Feature Learning 2011}, \bibinfo{year}{2011}.
%Type = Inproceedings
\bibitem[{LeCun and Cortes(2005)}]{LeCun2005TheMD}
\bibinfo{author}{Y.~LeCun}, \bibinfo{author}{C.~Cortes},
\newblock \bibinfo{title}{The mnist database of handwritten digits},
\newblock \bibinfo{year}{2005}.
%Type = Article
\bibitem[{Xiao et~al.(2017)Xiao, Rasul, and
  Vollgraf}]{DBLP:journals/corr/abs-1708-07747}
\bibinfo{author}{H.~Xiao}, \bibinfo{author}{K.~Rasul},
  \bibinfo{author}{R.~Vollgraf},
\newblock \bibinfo{title}{Fashion-mnist: a novel image dataset for benchmarking
  machine learning algorithms},
\newblock \bibinfo{journal}{CoRR} \bibinfo{volume}{abs/1708.07747}
  (\bibinfo{year}{2017}). \URLprefix \url{http://arxiv.org/abs/1708.07747}.
  \href{http://arxiv.org/abs/1708.07747}{{\tt arXiv:1708.07747}}.
%Type = Techreport
\bibitem[{Bulatov(2011)}]{nmnist}
\bibinfo{author}{Y.~Bulatov}, \bibinfo{title}{Notmnist dataset},
  \bibinfo{type}{Technical Report}, \bibinfo{year}{2011}.
%Type = Article
\bibitem[{Serr{\`{a}} et~al.(2018)Serr{\`{a}}, Sur{\'{\i}}s, Miron, and
  Karatzoglou}]{DBLP:journals/corr/abs-1801-01423}
\bibinfo{author}{J.~Serr{\`{a}}}, \bibinfo{author}{D.~Sur{\'{\i}}s},
  \bibinfo{author}{M.~Miron}, \bibinfo{author}{A.~Karatzoglou},
\newblock \bibinfo{title}{Overcoming catastrophic forgetting with hard
  attention to the task},
\newblock \bibinfo{journal}{CoRR} \bibinfo{volume}{abs/1801.01423}
  (\bibinfo{year}{2018}). \URLprefix \url{http://arxiv.org/abs/1801.01423}.
  \href{http://arxiv.org/abs/1801.01423}{{\tt arXiv:1801.01423}}.
%Type = Inproceedings
\bibitem[{Ke et~al.(2020)Ke, Liu, and Huang}]{10.5555/3495724.3497277}
\bibinfo{author}{Z.~Ke}, \bibinfo{author}{B.~Liu}, \bibinfo{author}{X.~Huang},
\newblock \bibinfo{title}{Continual learning of a mixed sequence of similar and
  dissimilar tasks},
\newblock in: \bibinfo{booktitle}{Proceedings of the 34th International
  Conference on Neural Information Processing Systems}, NIPS'20,
  \bibinfo{publisher}{Curran Associates Inc.}, \bibinfo{address}{Red Hook, NY,
  USA}, \bibinfo{year}{2020}.
%Type = Article
\bibitem[{Kingma and Ba(2014)}]{adam}
\bibinfo{author}{D.~Kingma}, \bibinfo{author}{J.~Ba},
\newblock \bibinfo{title}{Adam: A method for stochastic optimization},
\newblock \bibinfo{journal}{International Conference on Learning
  Representations}  (\bibinfo{year}{2014}).
%Type = Article
\bibitem[{Ruder(2016)}]{ruder2016overview}
\bibinfo{author}{S.~Ruder},
\newblock \bibinfo{title}{An overview of gradient descent optimization
  algorithms},
\newblock \bibinfo{journal}{arXiv preprint arXiv:1609.04747}
  (\bibinfo{year}{2016}).
%Type = Article
\bibitem[{Ebrahimi et~al.(2019)Ebrahimi, Elhoseiny, Darrell, and
  Rohrbach}]{DBLP:journals/corr/abs-1906-02425}
\bibinfo{author}{S.~Ebrahimi}, \bibinfo{author}{M.~Elhoseiny},
  \bibinfo{author}{T.~Darrell}, \bibinfo{author}{M.~Rohrbach},
\newblock \bibinfo{title}{Uncertainty-guided continual learning with bayesian
  neural networks},
\newblock \bibinfo{journal}{CoRR} \bibinfo{volume}{abs/1906.02425}
  (\bibinfo{year}{2019}). \URLprefix \url{http://arxiv.org/abs/1906.02425}.
  \href{http://arxiv.org/abs/1906.02425}{{\tt arXiv:1906.02425}}.
%Type = Article
\bibitem[{Musgrave et~al.(2020)Musgrave, Belongie, and
  Lim}]{DBLP:journals/corr/abs-2003-08505}
\bibinfo{author}{K.~Musgrave}, \bibinfo{author}{S.~J. Belongie},
  \bibinfo{author}{S.~Lim},
\newblock \bibinfo{title}{A metric learning reality check},
\newblock \bibinfo{journal}{CoRR} \bibinfo{volume}{abs/2003.08505}
  (\bibinfo{year}{2020}). \URLprefix \url{https://arxiv.org/abs/2003.08505}.
  \href{http://arxiv.org/abs/2003.08505}{{\tt arXiv:2003.08505}}.
%Type = Article
\bibitem[{Melis et~al.(2017)Melis, Dyer, and
  Blunsom}]{DBLP:journals/corr/MelisDB17}
\bibinfo{author}{G.~Melis}, \bibinfo{author}{C.~Dyer},
  \bibinfo{author}{P.~Blunsom},
\newblock \bibinfo{title}{On the state of the art of evaluation in neural
  language models},
\newblock \bibinfo{journal}{CoRR} \bibinfo{volume}{abs/1707.05589}
  (\bibinfo{year}{2017}). \URLprefix \url{http://arxiv.org/abs/1707.05589}.
  \href{http://arxiv.org/abs/1707.05589}{{\tt arXiv:1707.05589}}.
%Type = Article
\bibitem[{Dacrema et~al.(2019)Dacrema, Cremonesi, and
  Jannach}]{DBLP:journals/corr/abs-1907-06902}
\bibinfo{author}{M.~F. Dacrema}, \bibinfo{author}{P.~Cremonesi},
  \bibinfo{author}{D.~Jannach},
\newblock \bibinfo{title}{Are we really making much progress? {A} worrying
  analysis of recent neural recommendation approaches},
\newblock \bibinfo{journal}{CoRR} \bibinfo{volume}{abs/1907.06902}
  (\bibinfo{year}{2019}). \URLprefix \url{http://arxiv.org/abs/1907.06902}.
  \href{http://arxiv.org/abs/1907.06902}{{\tt arXiv:1907.06902}}.
%Type = Inproceedings
\bibitem[{Koch et~al.(2021)Koch, Denton, Hanna, and Foster}]{2021_3b8a6142}
\bibinfo{author}{B.~Koch}, \bibinfo{author}{E.~Denton},
  \bibinfo{author}{A.~Hanna}, \bibinfo{author}{J.~G. Foster},
\newblock \bibinfo{title}{Reduced, reused and recycled: The life of a dataset
  in machine learning research},
\newblock in: \bibinfo{editor}{J.~Vanschoren}, \bibinfo{editor}{S.~Yeung}
  (Eds.), \bibinfo{booktitle}{Proceedings of the Neural Information Processing
  Systems Track on Datasets and Benchmarks}, volume~\bibinfo{volume}{1},
  \bibinfo{publisher}{Curran}, \bibinfo{year}{2021}.
%Type = Misc
\bibitem[{Farquhar and Gal(2019)}]{farquhar2019robust}
\bibinfo{author}{S.~Farquhar}, \bibinfo{author}{Y.~Gal},
  \bibinfo{title}{Towards robust evaluations of continual learning},
  \bibinfo{year}{2019}. \href{http://arxiv.org/abs/1805.09733}{{\tt
  arXiv:1805.09733}}.
%Type = Article
\bibitem[{Chaudhry et~al.(2018)Chaudhry, Dokania, Ajanthan, and
  Torr}]{DBLP:journals/corr/abs-1801-10112}
\bibinfo{author}{A.~Chaudhry}, \bibinfo{author}{P.~K. Dokania},
  \bibinfo{author}{T.~Ajanthan}, \bibinfo{author}{P.~H.~S. Torr},
\newblock \bibinfo{title}{Riemannian walk for incremental learning:
  Understanding forgetting and intransigence},
\newblock \bibinfo{journal}{CoRR} \bibinfo{volume}{abs/1801.10112}
  (\bibinfo{year}{2018}). \URLprefix \url{http://arxiv.org/abs/1801.10112}.
  \href{http://arxiv.org/abs/1801.10112}{{\tt arXiv:1801.10112}}.
%Type = Article
\bibitem[{Mehta et~al.(2021)Mehta, Patil, Chandar, and
  Strubell}]{DBLP:journals/corr/abs-2112-09153}
\bibinfo{author}{S.~V. Mehta}, \bibinfo{author}{D.~Patil},
  \bibinfo{author}{S.~Chandar}, \bibinfo{author}{E.~Strubell},
\newblock \bibinfo{title}{An empirical investigation of the role of
  pre-training in lifelong learning},
\newblock \bibinfo{journal}{CoRR} \bibinfo{volume}{abs/2112.09153}
  (\bibinfo{year}{2021}). \URLprefix \url{https://arxiv.org/abs/2112.09153}.
  \href{http://arxiv.org/abs/2112.09153}{{\tt arXiv:2112.09153}}.
%Type = Misc
\bibitem[{Mirzadeh et~al.(2022)Mirzadeh, Chaudhry, Yin, Nguyen, Pascanu, Gorur,
  and Farajtabar}]{mirzadeh2022architecture}
\bibinfo{author}{S.~I. Mirzadeh}, \bibinfo{author}{A.~Chaudhry},
  \bibinfo{author}{D.~Yin}, \bibinfo{author}{T.~Nguyen},
  \bibinfo{author}{R.~Pascanu}, \bibinfo{author}{D.~Gorur},
  \bibinfo{author}{M.~Farajtabar}, \bibinfo{title}{Architecture matters in
  continual learning}, \bibinfo{year}{2022}.
  \href{http://arxiv.org/abs/2202.00275}{{\tt arXiv:2202.00275}}.
%Type = Article
\bibitem[{Wang et~al.(2022)Wang, Noll, Srinivasan, Gagnon-Bartsch, Kim, and
  Rao}]{doi:10.34133/2022/9807590}
\bibinfo{author}{N.~C. Wang}, \bibinfo{author}{D.~C. Noll},
  \bibinfo{author}{A.~Srinivasan}, \bibinfo{author}{J.~Gagnon-Bartsch},
  \bibinfo{author}{M.~M. Kim}, \bibinfo{author}{A.~Rao},
\newblock \bibinfo{title}{Simulated mri artifacts: Testing machine learning
  failure modes},
\newblock \bibinfo{journal}{BME Frontiers} \bibinfo{volume}{2022}
  (\bibinfo{year}{2022}). \URLprefix
  \url{https://spj.science.org/doi/abs/10.34133/2022/9807590}.
  \DOIprefix\doi{10.34133/2022/9807590}.
  \href{http://arxiv.org/abs/https://spj.science.org/doi/pdf/10.34133/2022/9807590}{{\tt
  arXiv:https://spj.science.org/doi/pdf/10.34133/2022/9807590}}.
%Type = Article
\bibitem[{Liu et~al.(2022)Liu, Glocker, McCradden, Ghassemi, Denniston, and
  Oakden-Rayner}]{Liu2022}
\bibinfo{author}{X.~Liu}, \bibinfo{author}{B.~Glocker}, \bibinfo{author}{M.~M.
  McCradden}, \bibinfo{author}{M.~Ghassemi}, \bibinfo{author}{A.~K. Denniston},
  \bibinfo{author}{L.~Oakden-Rayner},
\newblock \bibinfo{title}{The medical algorithmic audit},
\newblock \bibinfo{journal}{The Lancet Digital Health} \bibinfo{volume}{4}
  (\bibinfo{year}{2022}) \bibinfo{pages}{e384--e397}. \URLprefix
  \url{https://doi.org/10.1016/s2589-7500(22)00003-6}.
  \DOIprefix\doi{10.1016/s2589-7500(22)00003-6}.
%Type = Article
\bibitem[{Pellegrini et~al.(2021)Pellegrini, Lomonaco, Graffieti, and
  Maltoni}]{DBLP:journals/corr/abs-2105-13127}
\bibinfo{author}{L.~Pellegrini}, \bibinfo{author}{V.~Lomonaco},
  \bibinfo{author}{G.~Graffieti}, \bibinfo{author}{D.~Maltoni},
\newblock \bibinfo{title}{Continual learning at the edge: Real-time training on
  smartphone devices},
\newblock \bibinfo{journal}{CoRR} \bibinfo{volume}{abs/2105.13127}
  (\bibinfo{year}{2021}). \URLprefix \url{https://arxiv.org/abs/2105.13127}.
  \href{http://arxiv.org/abs/2105.13127}{{\tt arXiv:2105.13127}}.
%Type = Misc
\bibitem[{Wang et~al.(2022)Wang, Fink, Gool, and Dai}]{wang2022continual}
\bibinfo{author}{Q.~Wang}, \bibinfo{author}{O.~Fink}, \bibinfo{author}{L.~V.
  Gool}, \bibinfo{author}{D.~Dai}, \bibinfo{title}{Continual test-time domain
  adaptation}, \bibinfo{year}{2022}.
  \href{http://arxiv.org/abs/2203.13591}{{\tt arXiv:2203.13591}}.
%Type = Misc
\bibitem[{Tang et~al.(2023)Tang, Qendro, Spathis, Kawsar, Mascolo, and
  Mathur}]{tang2023practical}
\bibinfo{author}{C.~I. Tang}, \bibinfo{author}{L.~Qendro},
  \bibinfo{author}{D.~Spathis}, \bibinfo{author}{F.~Kawsar},
  \bibinfo{author}{C.~Mascolo}, \bibinfo{author}{A.~Mathur},
  \bibinfo{title}{Practical self-supervised continual learning with continual
  fine-tuning}, \bibinfo{year}{2023}.
  \href{http://arxiv.org/abs/2303.17235}{{\tt arXiv:2303.17235}}.
%Type = Book
\bibitem[{Huyen(2022)}]{huyen2022designing}
\bibinfo{author}{C.~Huyen}, \bibinfo{title}{Designing Machine Learning Systems:
  An Iterative Process for Production-ready Applications},
  \bibinfo{publisher}{O'Reilly Media, Incorporated}, \bibinfo{year}{2022}.
  \URLprefix \url{https://books.google.pl/books?id=BAy\_zgEACAAJ}.
%Type = Inproceedings
\bibitem[{Kohavi and Thomke(2017)}]{Kohavi2017TheSP}
\bibinfo{author}{R.~Kohavi}, \bibinfo{author}{S.~Thomke},
\newblock \bibinfo{title}{The surprising power of online experiments},
\newblock \bibinfo{year}{2017}.
%Type = Misc
\bibitem[{Semola et~al.(2022)Semola, Lomonaco, and
  Bacciu}]{semola2022continuallearningasaservice}
\bibinfo{author}{R.~Semola}, \bibinfo{author}{V.~Lomonaco},
  \bibinfo{author}{D.~Bacciu}, \bibinfo{title}{Continual-learning-as-a-service
  (claas): On-demand efficient adaptation of predictive models},
  \bibinfo{year}{2022}. \href{http://arxiv.org/abs/2206.06957}{{\tt
  arXiv:2206.06957}}.

\end{thebibliography}

\end{document}